# A Comprehensive Inference-Time Augmentation Framework in Physiological Signals: Application to PPG-Based AF Detection

**Davood Fattahi[1], Runze Yan[1], Saurabh Kataria[1], Zhaoliang Chen[2], Xiao Hu[1,3,4*]**

[1] Nell Hodgson Woodruff School of Nursing, Emory University, Atlanta, GA, USA
[2]Department of Computer Science, Emory University, Atlanta, GA, USA
[3] Wallace H. Coulter Department of Biomedical Engineering, Georgia Institute of Technology, Atlanta, GA, USA
[4]Department of Biomedical Informatics, Emory University School of Medicine, Atlanta, GA, USA

*Author to whom any correspondence should be addressed.
E-mail: xiao.hu@emory.edu

## Abstract

**Objective:** Accurate classification of physiological signals in real-world deployments is challenged by sensor noise, motion artifacts, and distribution shifts between training and deployment data. Inference-time augmentation (ITA), which applies augmentations during inference rather than retraining, offers a simple, model-agnostic mechanism to improve robustness. However, ITA application to physiological signals has remained narrow in scope, relying on limited augmentation methods with fixed, unoptimized parameters. This work proposes a unified ITA framework to address that gap.
**Approach:** The framework incorporates 13 augmentation methods spanning time-domain, amplitude-domain, frequency-domain, and artifact-injection transformations, with hyperparameters systematically optimized via Bayesian optimization. We evaluate the framework on atrial fibrillation (AF) detection from 30-second photoplethysmography (PPG) signals using two deep learning architectures, GPT-PPG (in two sizes) and ResNet, across five datasets comprising more than 400 patients and ~9,800 hours of PPG recording. Two evaluation strategies are assessed: standard ITA applied to all inputs, and selective ITA applied to initially positive predictions.
**Main results:** Standard ITA consistently improved the area under the receiver operating characteristic curve (AUROC, up to 8.5% for GPT-PPG and 0.7% for ResNet) and the area under the precision-recall curve (AUPRC, up to 10.6% for GPT-PPG and 0.8% for ResNet) across all model–dataset combinations. Selective ITA further reduced average FPR by up to 4.4% (GPT-PPG) and 1.3% (ResNet) on non-AF PPG datasets.
**Significance:** These findings establish ITA as a practical, model-agnostic approach for improving the reliability of PPG-based AF classification in deployment settings where retraining is not feasible, with broader applicability to physiological signal analysis.

## 1. Introduction

Inference-Time Augmentation (ITA) is a data augmentation strategy applied during model inference, such as when performing model evaluation or deployment. Unlike traditional training-time augmentation that expands and diversifies the training dataset to improve generalization, ITA generates multiple augmented variants of an input sample at inference time, propagates them through a trained model, and aggregates their predictions. This procedure acts as an inference-stage ensemble and, by simulating realistic variations of the input, can enhance prediction accuracy, calibration, and robustness to small perturbations and distribution shifts between training and deployment data, particularly in the presence of noisy or out-of-distribution inputs [1]. A key advantage of ITA compared to traditional augmentation is that it improves performance without any retraining, keeping the underlying model parameters unchanged. ITA is also model-agnostic and can be integrated into any predictive pipeline capable of processing different input variants. [1].

Tracing ITA back to its roots, traditional data augmentation is typically used to expand and diversify the training set during model-learning phase, especially when available data are limited [2, 3]. Among various data types, physiological waveforms present unique challenges and opportunities for augmentation. Unlike images or text, they are usually one-dimensional time-series data with temporal dependencies, physiological rhythms, and often cycle-specific patterns. As such, augmentation techniques require careful design to preserve signal integrity while introducing controlled variability [4]. Signal augmentation involves transformations applied to raw or processed time-series data to simulate realistic variability. These transformations may be in the time domain (e.g., time warping, jittering), the amplitude domain (e.g., amplitude scaling or morphing), or more abstract feature spaces (e.g., frequency or latent domain). Techniques may be deterministic or stochastic, handcrafted or learned, and can be applied globally or locally (e.g., cycle-wise). The applications of signal augmentation are widespread and impactful. It has been reported frequently in the literature applied on different physiological signals including electrocardiography (ECG) [5-13], Electroencephalography (EEG) [5, 14-16], electromyography (EMG) [4, 17], electrooculography (EOG) [4], and accelerometers [9, 18]. From the perspective of augmentation's objectives, in supervised learning augmentation can mitigate overfitting, improve robustness to unseen data, and increase model accuracy, especially when labeled data is limited. In self-supervised and contrastive learning, augmentations are essential to generate multiple "views" of the same underlying sample. In real-world deployments such as wearable health monitoring or remote sensing, augmentation can simulate sensor noise, motion artifacts, or inter-subject variability, making models more resilient to operational conditions.

Despite this widespread use of traditional augmentation during training, the application of augmentation at inference time remains comparatively underexplored, especially for physiological signals. ITA has gained momentum primarily in image classification using deep learning models [2, 19-21], but its adoption in physiological signal analysis is still limited. Existing ITA studies for physiological data tend to be narrow in scope: they typically rely on simple augmentation methods with fixed parameters [10], employ approaches that are not fully training-free during inference, or do not investigate the role of choosing appropriate augmentation parameters [18]. To the best of our knowledge, no prior work has examined a comprehensive suite of augmentation techniques within an ITA framework while systematically optimizing their parameters for physiological signal classification.

In this study, we address these gaps by examining 13 augmentation methods, encompassing both widely used transformations and several novel signal-level perturbations specifically designed for physiological waveforms. We incorporate them into a unified ITA framework to evaluate its feasibility and effectiveness for classification tasks. Furthermore, we propose a systematic hyperparameter optimization procedure to identify the most effective combination of augmentation methods and parameter settings for inference-time deployment.

We evaluate the framework on the task of atrial fibrillation (AF) detection using single-channel photoplethysmography (PPG) signals gathered from multiple independent datasets. In order to evaluate the impact of ITA, two pretrained deep learning architectures are employed: GPT-PPG, explored in two different model sizes, and ResNet34 as a representative simpler model. Performance is assessed using AUROC and AUPRC, enabling a rigorous comparison of ITA's effects on model performance across different datasets, model capacities, and augmentation configurations.

In addition to analyzing overall detection performance, we investigate the utility of a selective ITA strategy, in which ITA is applied only to initially positive predictions, for reducing false positive detections. This evaluation incorporates several non-AF PPG datasets to assess the ability of ITA to suppress spurious AF classifications.

The remainder of the manuscript is structured as follows. Section 2 introduces the augmentation methods used in this study, together with their theoretical foundations and characteristics relevant to physiological signals. Section 3 presents the overall ITA framework and the associated hyperparameter optimization strategy. Section 4 evaluates the proposed framework on the task of PPG-based AF detection, and the corresponding results and analyses are discussed in Section 5. Finally, Section 6 concludes the manuscript.

## 2. Physiological Signals Augmentation Methods

In this section, we discuss different types of augmentation methods applicable to physiological signal ITA. Augmentation methods for physiological signals can be conceptually divided into two broad categories. The first category consists of techniques that introduce non-informative or undesired components into the signal, such as noise, artifacts, baseline drift, or powerline interference or distortions. The purpose of these augmentations is to improve the robustness of the prediction to a range of plausible artifact realizations and reduces the likelihood that the final prediction is dominated by a particular form of signal degradation present in the original measurement [22]. The second category includes augmentations that apply controlled and physiologically plausible variations to the informative structure of the signal, such as subtle changes in timing, amplitude, or morphology. These methods aim to improve the generalizability of the prediction by introducing natural inter-subject, intra-subject, or device-related variability without destroying clinically meaningful features.

Augmentation methods for physiological signals must balance the introduction of realistic variability with preservation of physiological integrity and fixed signal length. Excessive distortion of temporal or morphological structure (e.g., PQRST complexes in ECG or systolic–dicrotic notches in PPG) can corrupt diagnostic features, while inadequate variation limits generalizability.

Sections 2.1–2.3 present the three proposed augmentation methods, each with full derivations. Section 2.4 summarizes standard augmentation techniques incorporated into the framework, with full details provided in Appendix.

### *2.1 Time Warping*

Time warping is an augmentation technique that stretches or compresses segments of a signal along the time axis, altering the local temporal dynamics. In physiological signals, natural variations in heart rate, breathing patterns, or reaction times can cause similar temporal distortions. Simulating time warping exposes models to these realistic variations, improving their ability to generalize across different physiological states and acquisition conditions.

Time warping methods can be broadly categorized into constant (linear) and dynamic (nonlinear) approaches. In constant time warping, a signal is uniformly stretched or compressed by applying a fixed scaling factor to its time axis. Although simple, constant warping has several limitations. It alters the total signal length, which may conflict with the fixed-size input requirements of many deep learning models. Moreover, it fails to represent localized temporal irregularities that are common in physiological signals.

In contrast, dynamic time warping applies a nonlinear transformation that varies across the signal. It allows local portions of the waveform to be stretched or compressed differently, producing more physiologically consistent variations. By ensuring that the mapping function transforms the normalized time interval $t \in [0,1]$ to a warped time $\tau \in [0,1]$, dynamic time warping maintains the original signal length and compatibility with model input specifications (see Figure 1).

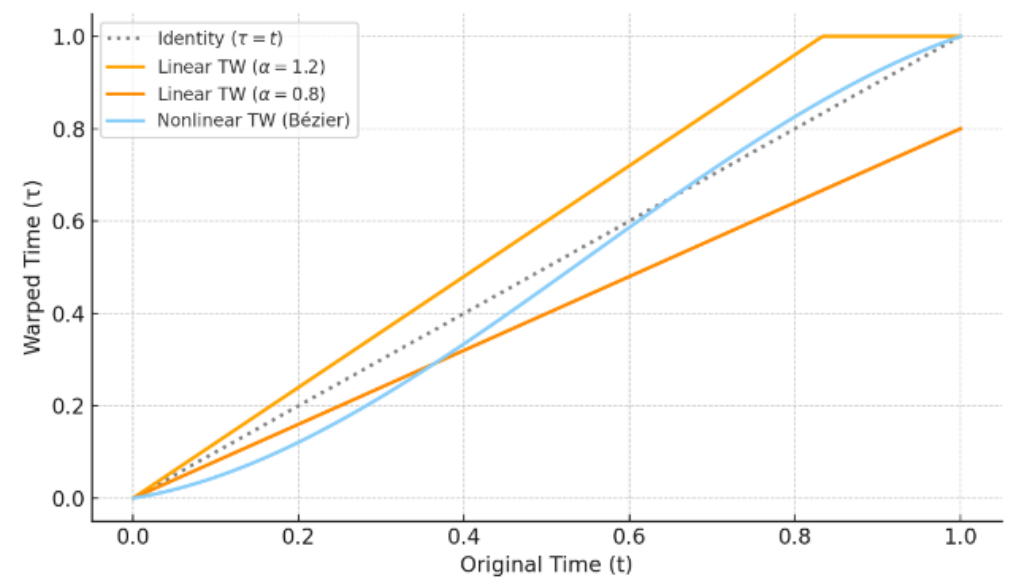

**Figure 1.** Linear time warping ($\tau = \alpha t$) changes the signal length. Nonlinear dynamic time warping addresses this issue. Horizontal axis: time in the input domain (before transformation); Vertical axis: time in the output domain (after transformation). Both axes are normalized by the signal length and are therefore dimensionless.

The dynamic time warping can be implemented by defining a time-mapping function $f(t)$ that transforms each normalized time point $t$ into a warped time $\tau = f(t)$, enabling localized stretching and compression of the signal. The augmented signal is obtained by resampling $x(t)$ at the warped indices,

$$\tilde{x}(t) = x(\tau) = x\big(f(t)\big). \qquad \text{Eq. (1)}$$

To preserve physiological integrity, $f(t)$ must satisfy the following conditions:

**Monotonicity**: $f(t)$ must be strictly increasing to preserve the chronological order of the signal samples. Non-monotonic warping would create temporal overlaps or reversals, corrupting the waveform.

**Smoothness**: $f(t)$ should vary continuously and gradually. Abrupt changes in the mapping would induce high-frequency artifacts and violate the smooth, band-limited nature of physiological signals.

**Centered around the identity**: The mapping should remain close to the identity line $\tau = t$. This ensures that the signal is perturbed realistically without being severely distorted.

**Fixed endpoints**: The mapping must satisfy $f(0) = 0$ and $f(1) = 1$ to preserve alignment at the start and end of the signal, ensuring that the full duration is retained.

According to above conditions, we suggest Bézier and Piecewise Cubic Hermite Interpolating Polynomial (PCHIP) functions as the mapping function of two different non-linear time-warping methods. Both follow a shared conceptual framework: they use a set of knots or control points whose randomized perturbations allow controlled variability, while the interpolation method (global polynomial in Bézier, local piecewise cubic in PCHIP) determines the smoothness and locality of the resulting mapping. Our nonlinear warping method follows three general steps that apply equally to both Bézier and PCHIP approaches:

**1. Initialization of control points:** A set of equally spaced control points $\tau_i \in [0,1]$is defined, with fixed endpoints $\tau_0 = 0$ and $\tau_K = 1$. For PCHIP, a corresponding set of $t_i \in [0,1]$is also defined to form knot pairs $(t_i, \tau_i)$.

**2. Adding randomness:** Uniform noise is added only to the interior points ($i = 1, \dots, K-1$).
Noise bounds are carefully selected to preserve strict monotonicity $\tau_0 < \tau_1 < \cdots < \tau_K$, (and similarly for the $t_i$ in PCHIP). Gaussian perturbations are avoided because they may create out-of-range values or violate this ordering.

**3. Constructing the warping function:** The final mapping $f(t)$ is constructed using either a Bézier curve or a PCHIP interpolant, providing smooth and physiologically consistent time distortions (see Figure 2). In the Bézier formulation, the mapping is obtained by evaluating the Bernstein polynomial expansion

$$f_{\text{Bézier}}(t) = \sum_{i=0}^{K} \binom{K}{i} (1-t)^{K-i} t^i \tau_i. \qquad \text{Eq. (2)}$$

But, PCHIP constructs the mapping using a sequence of knots $\{(t_i, \tau_i)\}_{i=0}^{K}$ , enabling more localized temporal variability. On each interval $[t_i, t_{i+1}]$, define $h_i = t_{i+1} - t_i$, and for each $t \in [t_i, t_{i+1}]$

$$s = \frac{t - t_i}{h_i}, \quad s \in [0,1], \qquad \text{Eq. (3)}$$

and let $m_i$ and $m_{i+1}$ denote the monotonicity-preserving slopes at $\tau_i$ and $\tau_{i+1}$. The PCHIP mapping on this interval is then

$$\begin{aligned} f_i(t) &= H_{00}(s)\,\tau_i + H_{10}(s)\,h_i m_i + \\ &\quad H_{01}(s)\,\tau_{i+1} + H_{11}(s)\,h_i m_{i+1}, \end{aligned} \qquad \text{Eq. (4)}$$

where the cubic Hermite basis functions are

$$\begin{aligned} H_{00}(s) &= 2s^3 - 3s^2 + 1, \\ H_{10}(s) &= s^3 - 2s^2 + s, \\ H_{01}(s) &= -2s^3 + 3s^2, \\ H_{11}(s) &= s^3 - s^2, \end{aligned} \qquad \text{Eq. (5)}$$

and the overall PCHIP function is

$$f_{\text{PCHIP}}(t) = \begin{cases} f_0(t), & t \in [t_0, t_1], \\ f_1(t), & t \in [t_1, t_2], \\ \vdots \\ f_{K-1}(t), & t \in [t_{K-1}, t_K]. \end{cases} \qquad \text{Eq. (6)}$$

Both approaches produce smooth, physiologically meaningful time variations but differ in the scale and locality of deformation: Bézier curves offer global control and smoothly varying distortions, whereas PCHIP provides sharper, segment-wise flexibility. Together, they offer complementary mechanisms for generating realistic nonlinear timing perturbations that enrich the augmented signal distribution while preserving the overall morphological integrity essential for accurate physiological inference (see Figure 2).

### *2.2 Amplitude Drift*

Amplitude drift is a gradual, low-frequency change in the baseline or overall magnitude of a physiological signal over time, often caused by factors such as sensor movement, changes in skin-electrode contact, temperature fluctuations, or slow physiological variations like respiration. In augmentation, simulating amplitude drift can help models learn to be invariant to such gradual shifts, ensuring that diagnostic features are not confounded by baseline fluctuations. This is especially important for long-duration recordings, where drift can accumulate and mask or distort clinically relevant patterns. By incorporating controlled amplitude drift during inference-time or training-time augmentation, the model is exposed to more realistic signal dynamics, enhancing its ability to maintain performance across diverse acquisition setups and patient conditions.

Dynamic amplitude drift can be implemented by constructing an additive amplitude deviator function $b(t)$, which is added to the original signal,

$$\tilde{x}(t) = x(t) + b(t). \qquad \text{Eq. (7)}$$

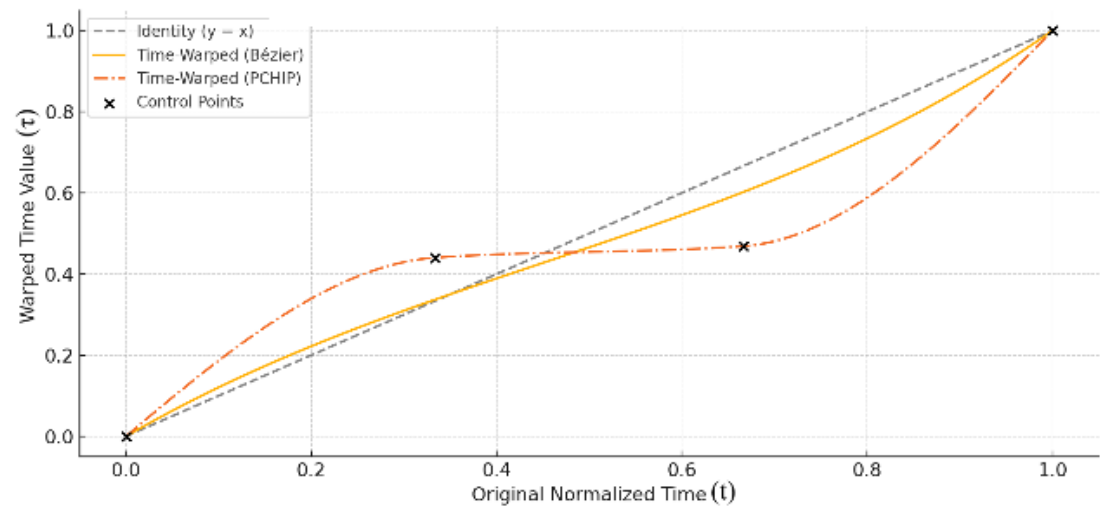

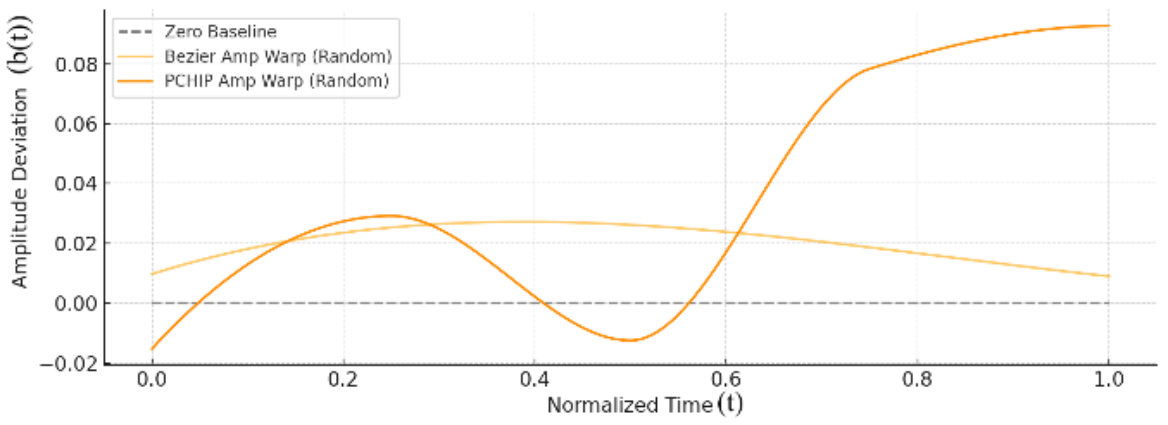

**Figure 2. Left:** Time warping mapping function using Bézier and PCHIP curves, using two random control points near y = x. Both the horizontal and vertical axes represent time normalized by the signal length and are therefore dimensionless. **Right:** $b(t)$ in amplitude drift and modulation, using Bézier and PCHIP curves. The horizontal axis represents time normalized by the signal length and is therefore dimensionless. The vertical axis represents the amplitude of $b(t)$, which is dimensionless when used for amplitude modulation and has the same units as the signal when used for amplitude drift.

This additive term plays the role of a dynamic baseline shift. It varies over time, typically in a slow and smooth manner, introducing naturalistic amplitude variability.

Amplitude drift can be categorized into linear and nonlinear forms, each representing different real-world baseline variations. Linear drift involves adding a gradually increasing or decreasing baseline, typically modelled as a straight slope across the signal, which can occur due to slow changes in sensor placement pressure or device calibration over time. Nonlinear drift, on the other hand, is modelled by adding a smoothly varying, nonlinear function, such as a sinusoid, polynomial curve, or spline, to the signal baseline. This form more closely resembles complex physiological or environmental influences, such as respiration-induced modulation, skin conductivity changes, or temperature effects. Including both linear and nonlinear drift in augmentation helps the model become resilient to a wide range of baseline shifts, ensuring it focuses on the underlying diagnostic features rather than being misled by gradual signal amplitude changes. In our augmentation framework, nonlinear amplitude drift is modelled using PCHIP and Bézier curves, both producing smooth, realistic baseline variations without abrupt discontinuities. However, unlike their use in time-warping, these curves are not constrained to be monotonic, as amplitude variations do not require strictly increasing behavior. Furthermore, the drift curves are constructed around the baseline $b(t) \approx 0$, rather than around an identity mapping (see Figure 2).

### *2.3 Amplitude Modulation*

Amplitude modulation (AM) in physiological signal augmentation involves varying the instantaneous amplitude of the signal over time, typically in a periodic or smoothly varying manner, while preserving its underlying frequency content. In real-world recordings, amplitude modulation can occur due to factors such as respiration-induced changes in blood volume (affecting PPG), muscle contractions influencing ECG amplitude, or changes in sensor contact pressure. By simulating AM during augmentation, the model is exposed to realistic fluctuations in signal strength that do not alter the fundamental waveform morphology. This helps the model learn to focus on shape- and pattern-based diagnostic features rather than being influenced by transient changes in amplitude. Mathematically, we can modulate the signal $x(t)$ by multiplying it with a curve which fluctuates around the value of 1. Therefore, we just need to change the Eq. (7) as follow

$$\tilde{x}(t) = x(t)\big(1 + b(t)\big), \quad \text{Eq. (8)}$$

where $b(t)$ can be implemented using Bézier and PCHIP curves to generate smooth, non-monotonic modulation patterns, with controllable variance and knot parameters to produce a diverse range of realistic amplitude fluctuation scenarios.

### *2.4. Other Augmentation Methods*

In addition to the three proposed methods above, the framework incorporates the following standard augmentation techniques. Additive noise superimposes synthetic noise onto the signal to simulate sensor noise and environmental interference. Burst augmentation introduces short, high-intensity transient disturbances to replicate sudden motion artifacts or brief sensor loss-of-contact events. Sine wave augmentation adds a periodic sinusoidal component to mimic powerline interference or cyclic physiological noise unrelated to the target measurement. Flip and invert augmentations reverse the signal along the time axis or negate its amplitude, respectively, to promote invariance to orientation and polarity differences across acquisition setups. Full mathematical descriptions and parameter definitions for each method are provided in Appendices A1-A4.

## 3. ITA Scheme and Hyper-Parameter Optimization

In our framework, ITA is implemented as a structured process that generates multiple augmented versions of each input signal and aggregates their predictions using probabilistic inference, either by averaging class likelihoods (soft voting) or by selecting the majority predicted class (hard voting). In our experiments, soft voting consistently yielded superior performance and was therefore adopted as the default aggregation strategy.

Given the large number of augmentation methods and their associated parameters, hyperparameter optimization (HPO) is employed to identify the most effective ITA configurations. Each method (e.g., noise addition, amplitude drift, time warping) is governed by parameters such as variance, number of control knots, frequency, and amplitude, forming a high-dimensional search space where each configuration represents a distinct augmentation strategy (see Table 1). To make the optimization tractable, parameters were grouped by augmentation method, assuming independence across groups. Each group was optimized separately while keeping the others disabled. The number of optimization iterations was scaled according to the complexity of the parameter space, methods with larger or continuous spaces (e.g., additive noise) received more trials, whereas simpler, discrete transformations (e.g., flipping or inversion) required fewer.

Most augmentation groups were optimized using Bayesian optimization with Gaussian process surrogates, which iteratively evaluates candidate configurations and updates a surrogate model to balance exploration and exploitation. For simple deterministic augmentations, exhaustive or grid-like search was sufficient. The optimization objective was defined as the validation performance (AUROC or AUPRC), and the loss function was set to $loss = 1 - metric$ to align with the minimization framework.

After obtaining the optimized hyperparameters for each augmentation group independently, a fusion strategy was required to apply all augmentations jointly. A naïve approach would be to apply each method separately and concatenate all resulting augmented versions; however, this would drastically increase the number of augmented variants to $\sum_k n_k$, where $n_k$ is the number of augmentations generated by the $k^{\text{th}}$method, leading to substantial computational cost. Instead, we adopted a hybrid fusion scheme, in which all augmentation methods are merged using their optimized parameters so that each augmented variant benefits simultaneously from multiple transformations. The only parameter not directly mergeable is the number of augmented variants, which appears in all groups; this was therefore set to the average of the optimized values across methods. This hybrid strategy provides a balanced trade-off between computational efficiency and the diversity of augmented samples.

## 4. Task-level Evaluation: PPG AF Detection

The proposed ITA framework is broadly applicable to a wide range of physiological signals, including ECG, PPG, PCG, EMG, EOG, and EEG, because its augmentation methods are formulated at the signal level and do not rely on modality-specific assumptions. Nonetheless, its benefits are expected to be more pronounced in signals that exhibit distinct morphological patterns, such as ECG and PPG. In these modalities, temporal and amplitude perturbations (e.g., time-warping, drift, and modulation) correspond to clinically meaningful variations in physiological parameters, making such augmentations particularly relevant and realistic.

**Table 1.** Different augmentation methods and their dedicated parameters

| Method | Parameters | Search Space |
|---|---|---|
| Additive Noise | Enable | True, False |
| | Number of augmented variants; | 1 – 15, integer |
| | Signal-to-noise ratio (dB); | 10 – 40, integer |
| | Noise color; | white, pink, brown, violet |
| | Bandpass filter low cutoff; | 0 – 0.4, real (Nyquist scaled) |
| | Bandpass filter high cutoff; | 0.5 – 1, real (Nyquist scaled) |
| | Noise distribution; | Gaussian, Uniform, Laplace, resampled |
| Bézier-Based Time Warping | Enable | True, False |
| | Number of augmented variants; | 1 – 15, integer |
| | Warping intensity; | 0.0 – 0.1, real |
| | Number of control knots; | 3 – 5, integer |
| Bézier-Based Amplitude Drift | Enable | True, False |
| | Number of augmented variants; | 1 – 15, integer |
| | Drift intensity; | 0.0 – 0.2, real |
| | Number of control knots; | 3 – 5, integer |
| Bézier-Based Amplitude Modulation | Enable | True, False |
| | Number of augmented variants; | 1 – 15, integer |
| | Modulation intensity; | 0.0 – 0.2, real |
| | Number of control knots; | 3 – 5, integer |
| PCHIP-Based Time Warping | Enable | True, False |
| | Number of augmented variants; | 1 – 15, integer |
| | Warping intensity; | 0.0 – 0.1, real |
| | Number of control knots; | 3 – 5, integer |
| PCHIP-Based Amplitude Drift | Enable | True, False |
| | Number of augmented variants; | 1 – 15, integer |
| | Drift intensity; | 0.0 – 0.2, real |
| | Number of control knots; | 3 – 5, integer |
| PCHIP-Based Amplitude Modulation | Enable | True, False |
| | Number of augmented variants; | 1 – 15, integer |
| | Modulation intensity; | 0.0 – 0.2, real |
| | Number of control knots; | 3 – 5, integer |
| Low-Frequency Noise Injection | Enable | True, False |
| | Number of augmented variants; | 1 – 15, integer |
| | Laplace noise amplitude; | 0 – 1, real |
| | Low-pass cut-off frequency; | 0.01 – 0.1, real (Nyquist scaled) |
| Sinusoidal Interference Injection | Enable | True, False |
| | Number of augmented variants; | 1 – 15, integer |
| | Sinusoidal wave amplitude; | 0 – 1, real |
| | Sinusoidal frequency; | 0.01 – 0.1, real (Nyquist scaled) |
| Burst Artifact Injection | Enable | True, False |
| | Number of augmented variants; | 1 – 15, integer |
| | Burst amplitude; | 0 – 1, real |
| | Number of bursts per signal; | 1 – 10, integer |
| | Burst noise frequency; | 0.01 – 0.1, real (Nyquist scaled) |
| Basic Deterministic Transformations | Reverse the signal in time (flipping); | True, False |
| | Reverse the signal polarity (inverting); | True, False |
| | Apply linear baseline drift | True, False |

To demonstrate the effectiveness and generality of all augmentation types within our ITA framework, we evaluate it on PPG signals for AF detection, as one of the most established yet challenging problems in the physiological signal analysis domain. PPG morphology is sensitive to cardiorespiratory dynamics, vascular compliance, and motion artifacts, making it an ideal modality for assessing the robustness and generalizability enhancements provided by ITA.

### *4.1 PPG-based AF Detection*

Atrial fibrillation (AF) is the most common sustained cardiac arrhythmia and a major contributor to stroke and cardiovascular morbidity, making reliable early detection essential. While ECG is the gold standard for diagnosing AF, its limited suitability for long-term, daily monitoring has led to increasing use of PPG signals, which are readily available from consumer wearable devices. However, PPG signals collected in free-living conditions are highly susceptible to noise, motion artifacts, and signal-quality fluctuations, making robust AF detection challenging [23-27].

These characteristics make AF detection from PPG an ideal testbed for evaluating ITA. PPG morphology is sensitive to subtle temporal and amplitude variations but is also easily corrupted by artifacts. Standard DNN models often degrade when confronted with such noise, especially when training labels contain uncertainty or when noisy samples are discarded during preprocessing.

ITA provides a principled alternative: rather than rejecting noisy inputs or relying on error-prone enhancement pipelines, multiple augmented versions of the test signal are generated, and their predictions are aggregated to improve robustness. Evaluating ITA on PPG for AF detection therefore allows us to demonstrate its ability to stabilize predictions, mitigate the effect of noise, and enhance model reliability in a clinically meaningful and challenging setting.

### *4.2 Datasets*

#### *4.2.1 UCSF Alarm Dataset:*

This dataset consisted of continuous bedside PPG recordings from 28,539 hospital patients. Bedside monitors automatically annotated rhythm events such as atrial fibrillation (AF) and premature ventricular contractions (PVC). For this study, merging PVC and normal sinus rhythm (NSR) into a single Non-AF category, the data is utilized in two classes for binary AF vs. Non-AF classification. PPG signals were segmented into non-overlapping 30-second windows, originally sampled at 240 Hz (7,200 points) and subsequently down-sampled to 80 Hz per second. Data were partitioned by patient ID into separate training and validation sets. The training set included 13,432 patients with 2,757,888 AF segments and 3,014,334 Non-AF segments; the validation set included 6,616 patients with 1,280,775 AF and 1,505,119 Non-AF segments [27]. Collection and use of the data were approved by Institutional Review Board (IRB) No. 14-13262.

#### *4.2.2 Stanford AF Dataset:*

This is a public dataset from Torres-Soto and Ashley [28], under Stanford University IRB protocol ID 35465, featuring data from wrist-worn devices in ambulatory settings. Originally with 25-second segments, we augmented these to 30 seconds and resampled to 80 Hz per second. It contains 52,911 AF and 80,620 Non-AF samples from 163 patients, including those with AF and healthy individuals [28].

#### *4.2.3 UCLA AF Dataset:*

This dataset was collected from routine use of bedside patient monitors in acute care units at the University of California Los Angeles medical center, and consists of fingertip PPG data from 126 hospital patients admitted between April 2010 and March 2013, with ages ranging from 18 to 95 years old. We formatted the continuous signals into 30-second, non-overlapping segments, down sampled to 80 Hz per second. This dataset includes 38,910 AF and 220,740 Non-AF segments, annotated by cardiac electrophysiologists [27]. Collection and use of the data were approved by IRB No. 10-000545.

#### *4.2.4 Simband AF Dataset:*

This dataset was gathered in Emory University using wrist-worn Samsung Simband devices from 98 ambulatory patients admitted between October 2015 and March 2016, aged between 18-89 years old. We processed the data into 30-second segments and down-sampled to 80 Hz per second. The dataset includes 348 AF and 506 Non-AF segments, reviewed and annotated by medical professionals [27, 29]. The recording and utilizing of the data were performed under IRB No. 00084629.

#### *4.2.5 Delta Dataset:*

This data is recorded from patients who recently had an acute stroke and are clinically being followed with 30-day extended ECG-patched based monitoring to detect AF (Approved by the Emory IRB, protocol 2025P010026). Enrolled patients are also provided with a MOTO 360 smartwatch to collect continuous PPG signals over approximately 30 days, in addition to the ECG patches. We use data from three participants who were confirmed to have AF events through ECG, showed by EM-019, EM-022, and EM-032 patient IDs, with 684-, 677-, and 681-hours duration and 1.95%, 6.06%, 15.26% AF burden, respectively. The data is segmented into 30-second non-overlapped pieces and resampled to 80 Hz per second.

### *4.2.6 Collection of Non-AF Datasets:*

This dataset consists of non-AF PPG signals collected from publicly available online sources. We initially screened 47 datasets and excluded those likely to contain AF or other cardiovascular abnormalities, including ICU recordings, post-stroke PPGs, post-surgical datasets, and datasets explicitly noting the presence of cardiovascular disease. From the remaining 17 datasets (selected based on a high likelihood of representing non-AF subjects and a minimum recording duration of 20 minutes per individual) we extracted approximately 4,500 hours of PPG data. These signals were preprocessed, segmented into 30-second windows, frequency-filtered, normalized, and resampled to 80 Hz. List of included datasets can be found in Appendix A5.

### *4.3 Models*

To evaluate the effectiveness of the proposed ITA framework, we selected two deep neural network (DNN)–based models that are recognized among the state-of-the-art in PPG-based AF detection. Although numerous classical (non-DNN) approaches have been published in this domain, we hypothesize that DNN models may benefit more substantially from ITA due to their higher model complexity and greater susceptibility to uncertainty and overfitting. Their large number of learnable parameters can lead to sharper and less stable decision boundaries, making them more responsive to augmentation strategies designed to enhance prediction robustness and generalizability.

**GPT-PPG:** This model has been recently introduced, and is a generative pre-trained transformer (GPT)-based foundation model specifically tailored to photoplethysmography (PPG) signals, enabling effective adaptation to various downstream biomedical tasks [30]. In this study we utilized the AF-detection tailored model in two 19 million and 85 million sizes, pre-trained and finetuned on UCSF dataset PPG segments with 40 Hz sample rate and 30-second duration. For the ITA phase, HPO was conducted on a randomly selected 5% subset of the same dataset.

**ResNet34:** Although this model was originally introduced for image recognition tasks, it has been successfully adapted for one-dimensional physiological signals, including PPG-based AF detection. Prior studies have demonstrated that residual architectures are particularly effective in PPG-based arrhythmia detection due to their ability to preserve waveform integrity while learning robust, multi-scale representations of rhythmic irregularity. In our ITA experiments, ResNet34 serves as a strong baseline DNN model against which the benefits of inference-time augmentation can be rigorously assessed [31, 32]. This model was pre-trained on UCSF dataset PPG segments with 40 Hz sample rate and 30-second duration, while for the ITA phase, HPO was conducted on the same subset used in GPT-PPG model (a randomly selected 5% subset of UCSF dataset).

## 5. Results and Discussion

### *5.1. Improvement in AF Detection*

Figure 3 illustrates the impact of the proposed ITA scheme on the performance of the three evaluated models, measured by AUROC and AUPRC across five PPG datasets. Each bar represents the percentage change in performance relative to the corresponding baseline model without ITA. Although the magnitude of improvement varies among models, datasets, and metrics, ITA consistently enhances performance, showing positive change in all cases and no degradation in any configuration.

To assess the effect of ITA across varying PPG signal qualities, we employed the Artifact Tolerance Curve (AT-curve) as an analytical tool. In this approach, data quality is gradually reduced at each step by imposing an upper limit on the quality index of the included samples. This allows evaluation of model performance across multiple subsets of data representing different quality levels. More information about the data quality calculation method can be found in [31].

Figure 4 demonstrate the AT-curve of AF detection in 4 different model-metric pairs for one of the datasets with wide range of signal quality index. An interesting observation is that ITA yields greater performance improvements when higher-quality PPG signals are included. This can be explained by the nature of ITA itself: the method intentionally introduces controlled perturbations to slightly distort the signal and reinforce the model's confidence in its class prediction. When the original PPG signals are clean, these perturbations generate additional, still-valid variations that help the model generalize better. In contrast, when the input signals are already heavily contaminated with artifacts, applying ITA adds further distortions that do not contribute to meaningful augmentation and therefore offer limited benefit.

The kernel density estimation (KDE) plot in Figure 5 illustrates the probability density distributions of distances from the decision boundary in GPT-PPG 19M model latent-space feature, comparing data before and after ITA for both AF and non-AF classes in Simband dataset. Each smooth curve represents how frequently particular distance values occur within a group, with the area under each curve normalized to one. By visualizing these continuous distributions rather than discrete counts, the plot highlights how ITA influences the separation between AF and non-AF features in the latent space. For instance, better inter-class separation and intra-class compactness after applying ITA indicates improved discriminability in this case.

To interpret how ITA affects the model's internal representations, we introduce a novel visualization approach named *Decision-Aligned Principal Component Analysis* (DA-PCA) that projects high-dimensional latent feature vectors into a two-dimensional space. The horizontal axis corresponds to the signed distance of each sample from the decision boundary (same as the horizontal axis used in Figure 5) while the vertical axis captures the first principal component of PCA in the subspace orthogonal to the decision axis, reflecting the dominant source of variance uninformative to the classification score. Together, these axes yield an interpretable view of latent-space structure that cleanly separates decision-relevant and decision-irrelevant variation.

Figure 6 presents DA-PCA projections before and after applying ITA, with AF and non-AF samples shown in distinct colors. In the pre-ITA panel, the two classes show considerable overlap, most noticeably for the cluster in the lower-left region of the panel. After ITA, the class distributions shift outward along the decision axis, with the overlap region substantially reduced, indicating that the aggregated prediction benefits from a stronger and more consistent discriminative margin. The post-ITA panel also shows a higher concentration of AF samples (red dots) on the correct side of the decision boundary, consistent with the improvement in true positive rate. Figures 5 and 6 together provide direct geometric evidence that ITA strengthens the model's discriminative structure in latent space.

Figure 7 illustrates the DA-PCA projections of latent-space features for four representative test samples from two classes, demonstrating both successful and unsuccessful cases in which ITA attempts to correct the predicted label. Although our classifier does not employ any KNN-based decision rule, we additionally highlight the KNN neighbourhoods to visualize the proximity of each test point to the surrounding training samples and to provide an intuitive sense of local decision uncertainty near the boundary. The plots show that ITA effectively shifts test features toward the correct class when the original sample lies close to the decision boundary, whereas its ability to correct the prediction diminishes when the sample is initially located far within the wrong decision region.

The positive effect of ITA on inter-class separation can also be observed in Figure 8, which show Pairwise Controlled Manifold Approximation Projection (PaCMAP) visualizations of the latent feature space [33]. So, two examples of features before the final linear classifier and output from the GPT encoder layer are illustrated. Both plots exhibit a relative improvement in class separation after applying ITA.

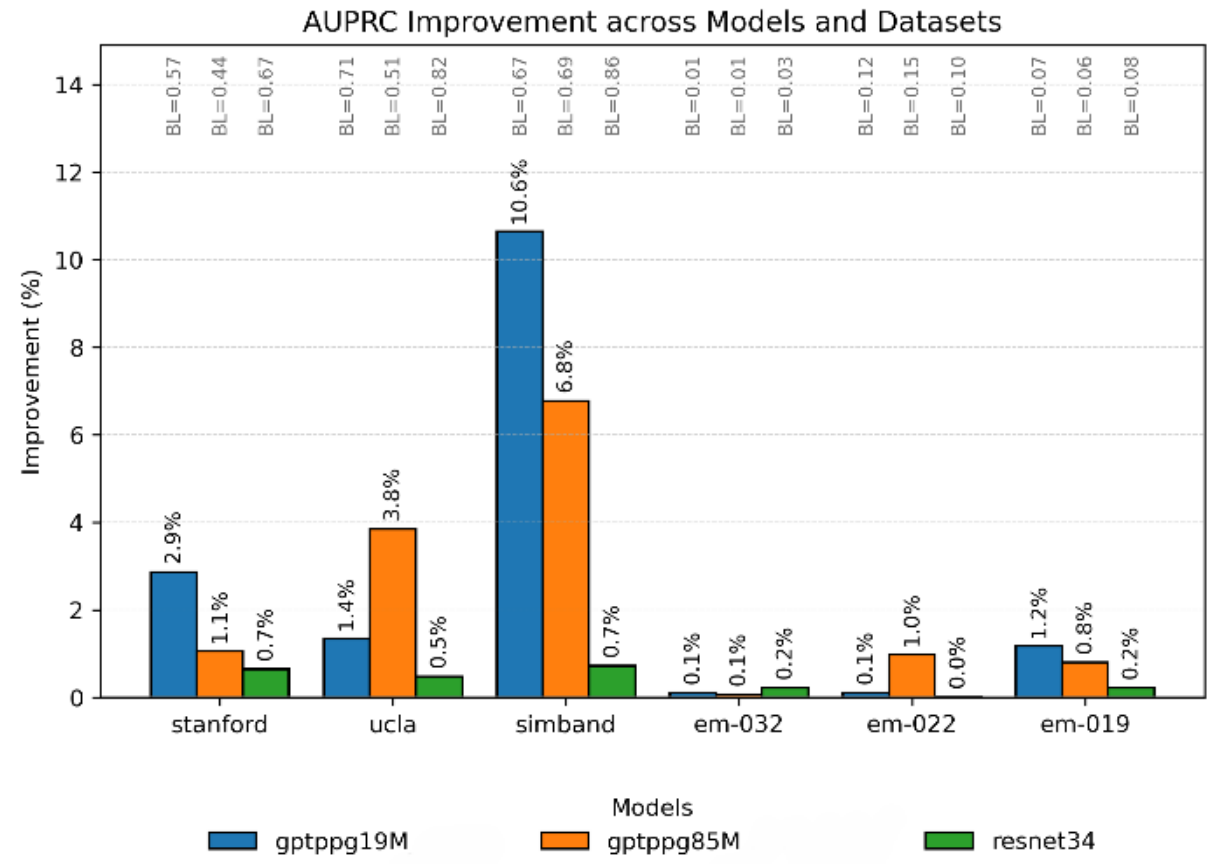

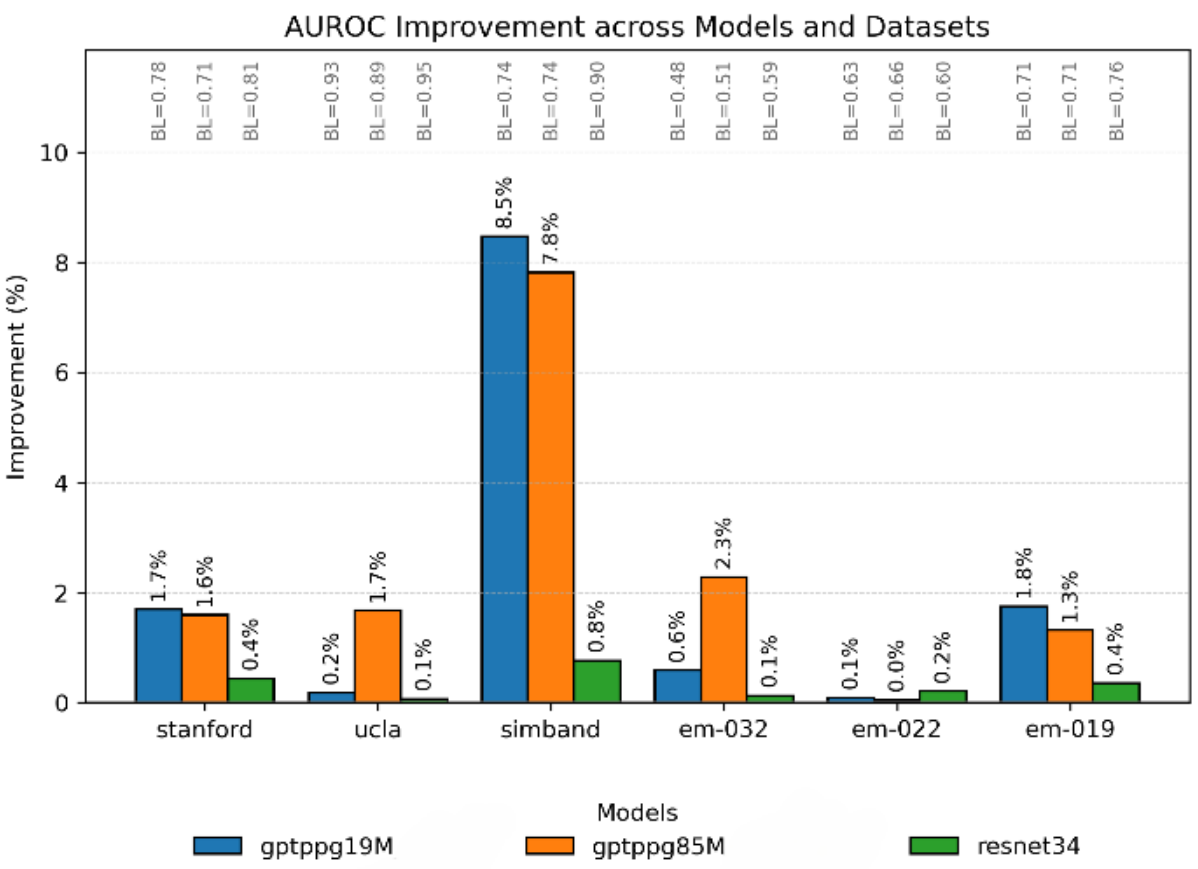

**Figure 3.** Percent of AUROC and AUROC improvement after applying ITA for GPT-PPG and ResNet34 models, across different datasets. BL: the metric value for the baseline model (without ITA).

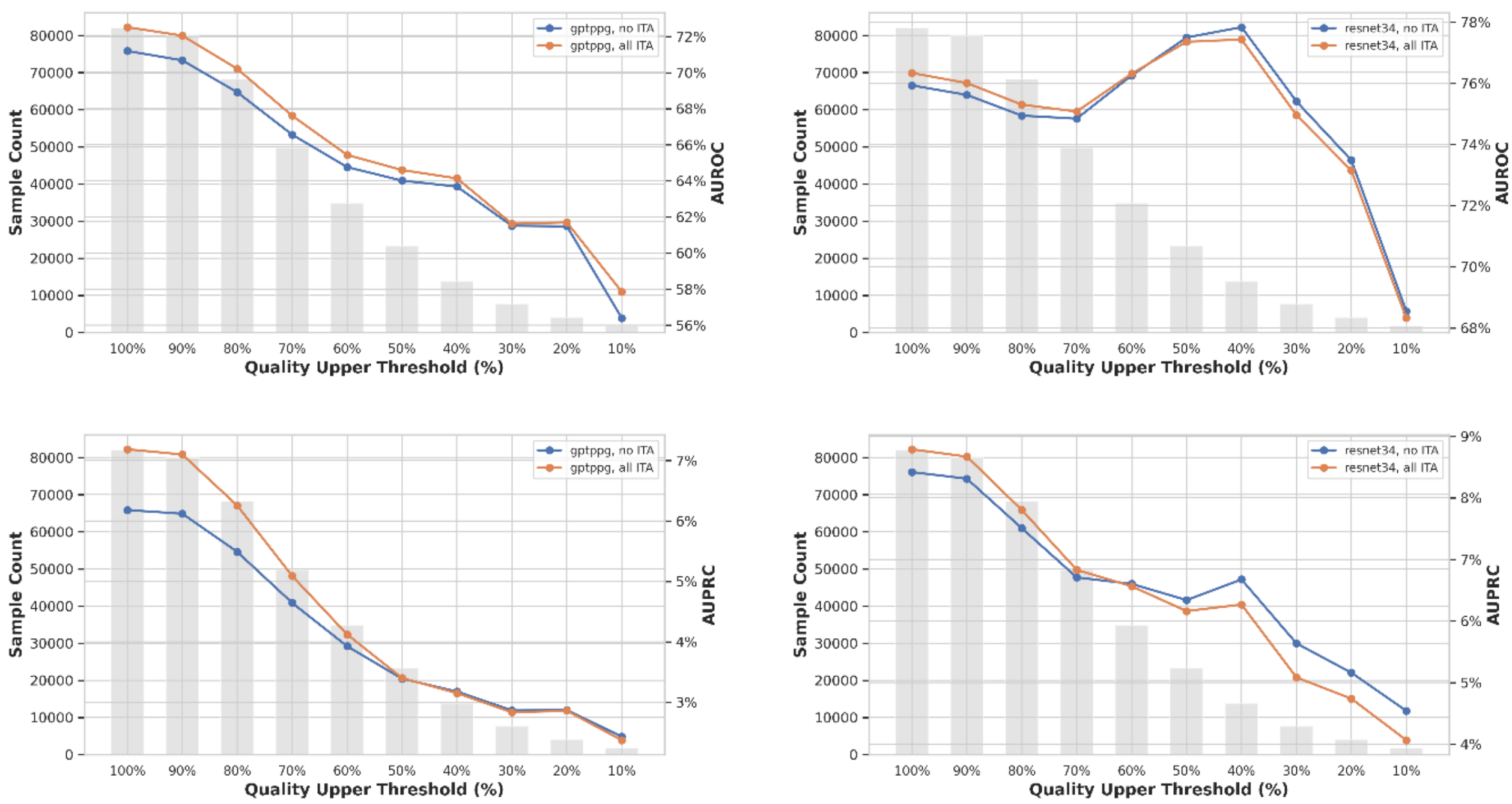

**Figure 4.** AT-curve of two models' performance for detecting AF using EM-32 PPG data; Up-left: GPT-PPG-85M model and AUROC performance metric, down-left: GPT-PPG-85M model and AUPRC performance metric, up-right: ResNet34 model and AUROC performance metric, down-right: ResNet34 model and AUPRC performance metric.

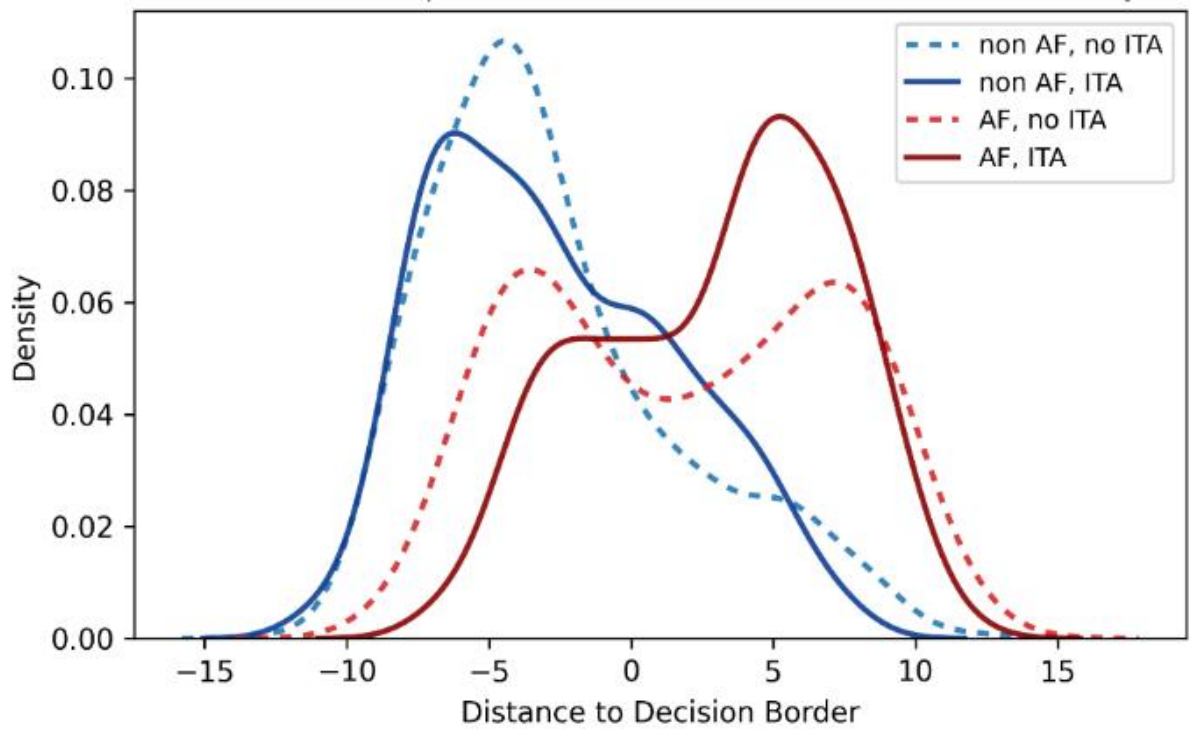

**Figure 5.** KDE of distances between samples and the decision boundary, before and after ITA. Features come from latent-space before the final linear classifier layer of GPT-PPG 19M model, and Simband dataset.

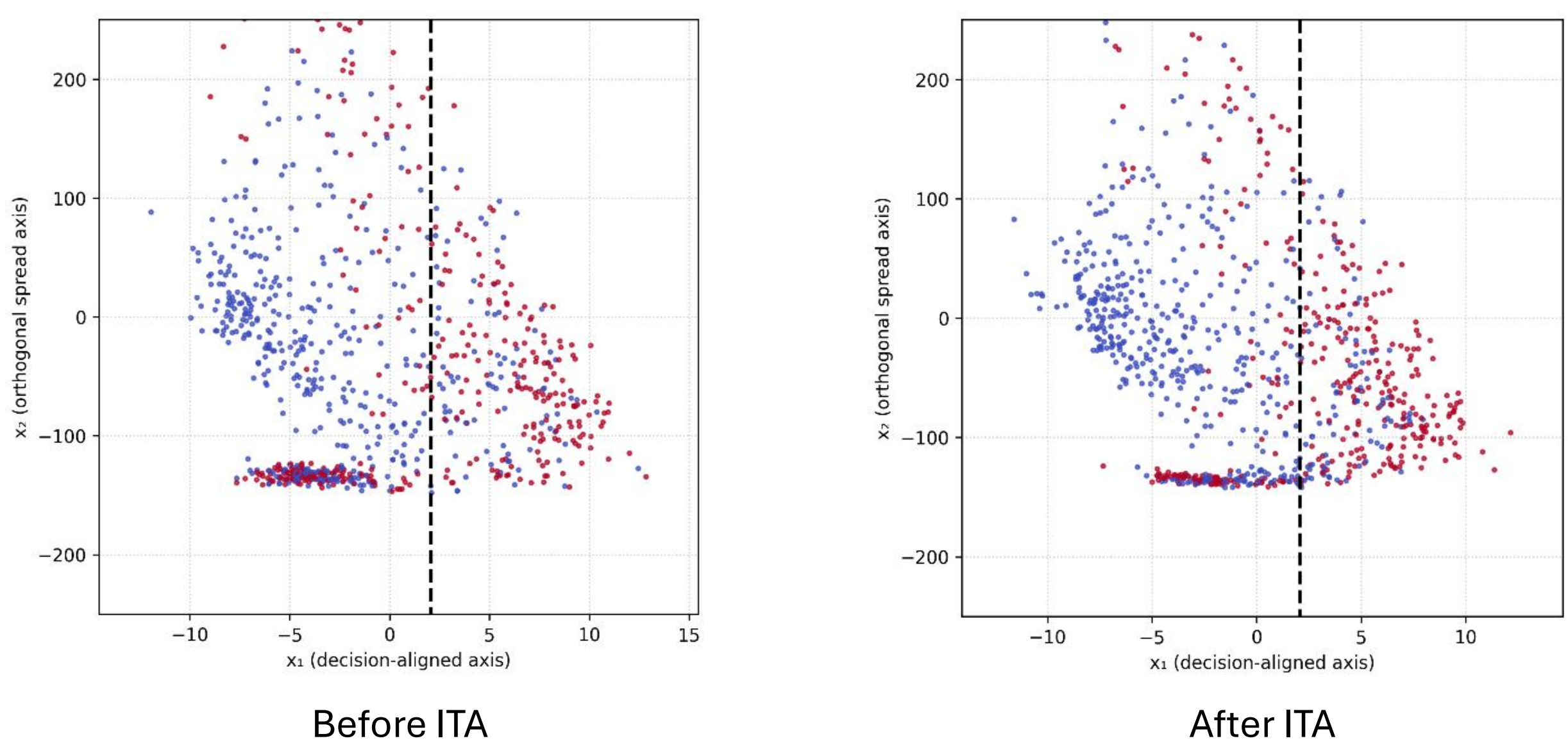

**Figure 6.** Decision-aligned PCA projection of latent-space features before the final linear classifier layer of GPT-PPG 19M model, and Simband dataset. Horizontal axis: the decision-aligned axis, distances between samples and the decision boundary. Vertical axis: first PCA component orthogonal to the decision-aligned axis. Blue: non-AF, red: AF.

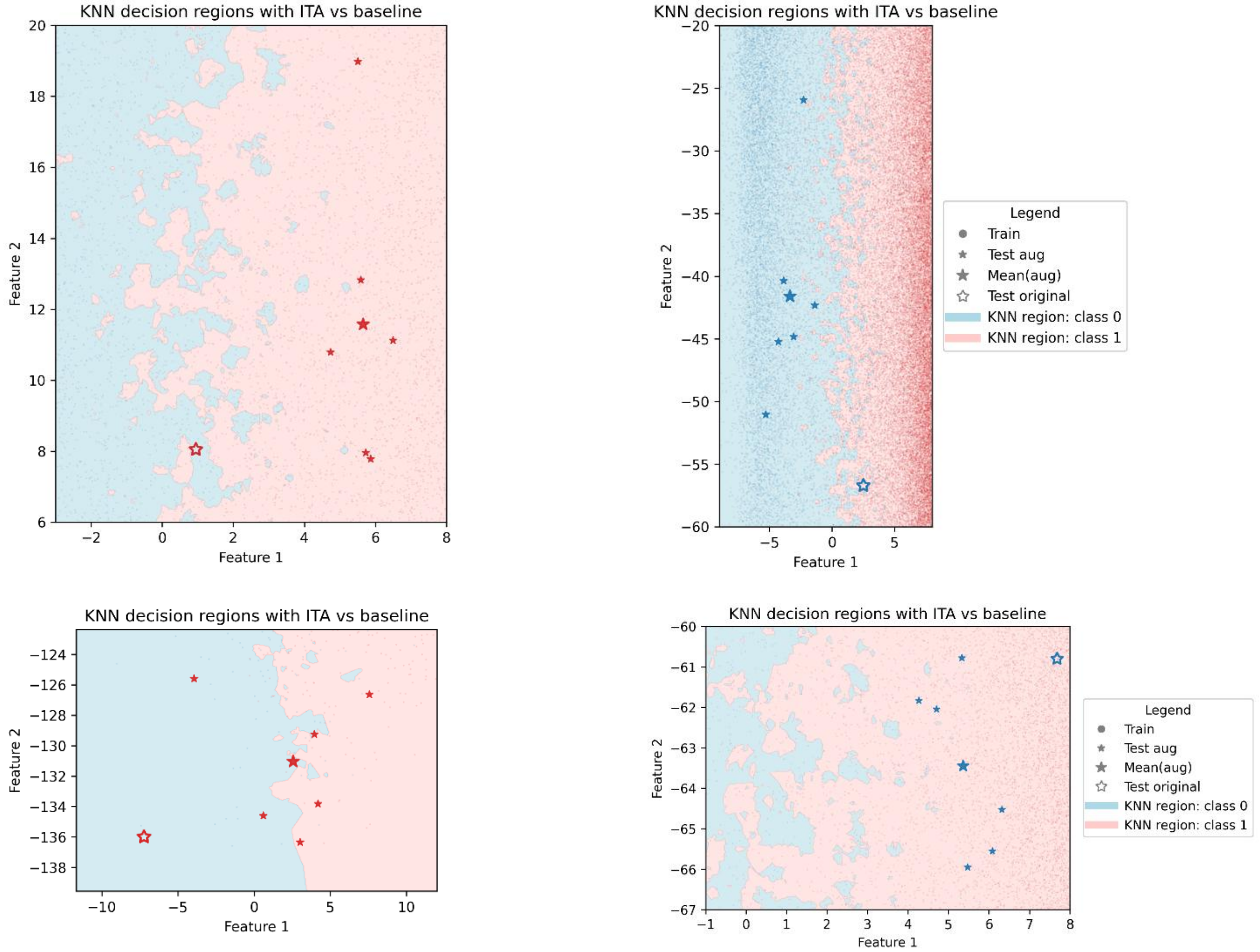

**Figure 7.** Example of successful (top) and unsuccessful (bottom) effect of ITA in changing misclassified samples. Blue: non-AF, red: AF.

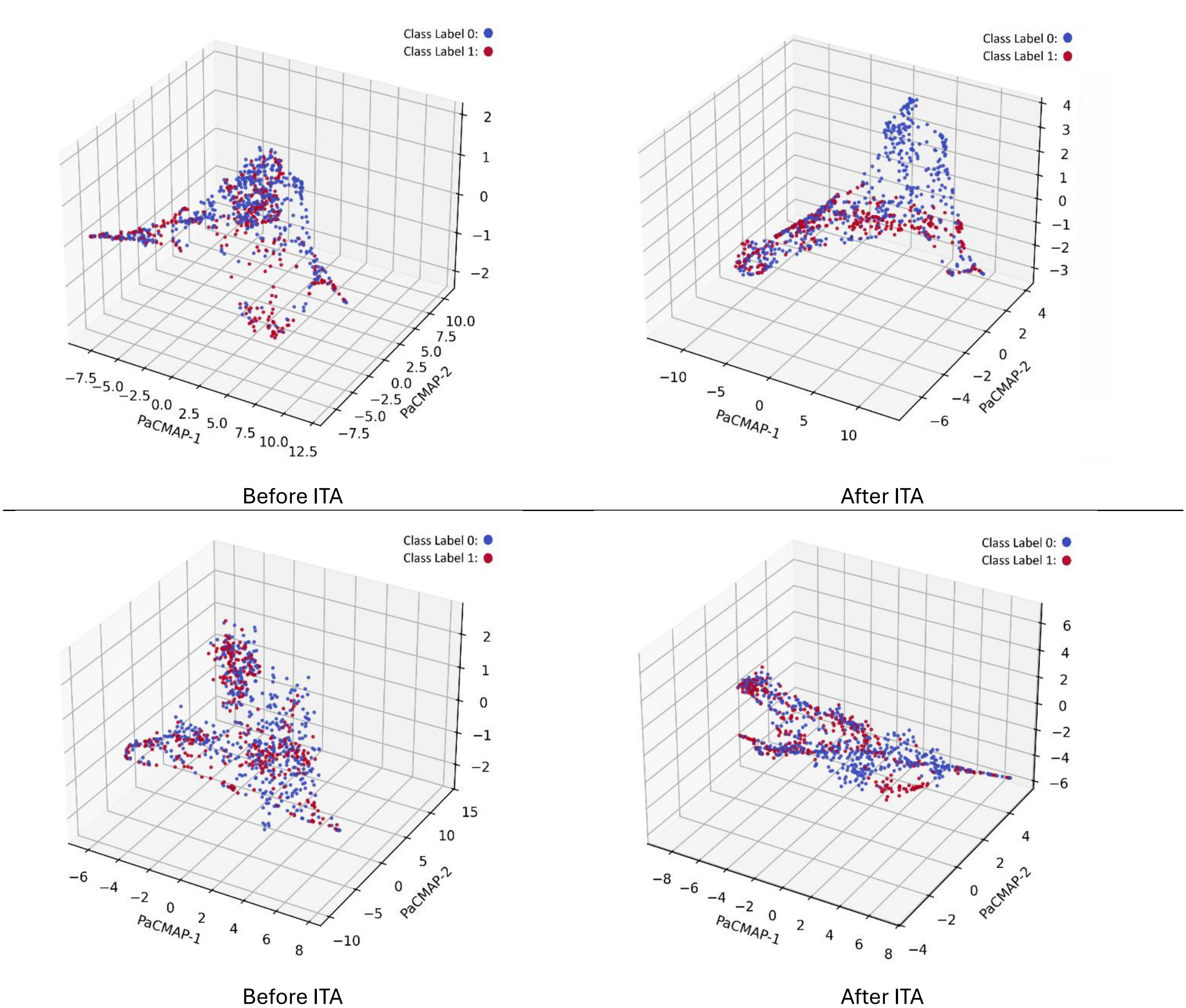

**Figure 8.** 3D PaCMAP plot of latent feature space of GPT-PPGG 19M model and Simband dataset, before and after ITA. Top: before the final linear classifier. Bottom: after GPT layer (GPT encoded features). Blue: non-AF, red: AF.

### *5.2 False Positive Ratio Reduction Using Selective ITA*

In many AF alarm mechanisms, the detection pipeline operates in two stages. The first stage is designed for candidate flagging, where the algorithm quickly identifies segments that potentially exhibit AF-like patterns. Because this step must operate efficiently and in real time, it typically prioritizes sensitivity over specificity, which can result in a relatively high number of false positives. To address this issue, a second-stage logic or sanity check is often applied to re-evaluate the initially flagged segments. This secondary stage may employ more computationally intensive analyses, stricter decision thresholds, second modality incorporation, or minimum duration criteria to confirm or reject AF candidates, thereby improving the reliability of the alarm system [34-36].

In this section, we investigate whether the proposed ITA framework can be utilized as the second stage of validation to reduce FPR in AF detection, when selectively applied on positive predicted samples.

The effectiveness of selective ITA has been reported in the literature for image data, where the selection is based on higher predictive uncertainty of a model [20]. However, in this study, we just utilize a simple selection criterion based on positive prediction of the model, aiming to reduce the FPR.

The key intuition is that a true AF segment should yield consistent predictions across its augmented variants, while a false alarm, originating from noise, motion artifacts, or irregular but non-AF rhythms, would exhibit more inconsistent outputs and thus be more likely down-weighted after soft voting. To evaluate this capability, we extended our analysis to include a collection of non-AF PPG datasets in addition to the AF datasets (refer to Section 4.2.6).

The results presented in Figure 9 demonstrate that the proposed ITA framework consistently lowers the false positive ratio across models and datasets. This highlights ITA's potential not only as a performance-enhancing mechanism but also as a built-in regularization tool for improving the reliability of AF detection systems in real-world conditions.

Employing ITA as a built-in sanity-check mechanism not only reduces false positives but can also enhance performance metrics for true AF cases. Figure 10 illustrates the improvement in AUROC and AUPRC obtained after applying ITA to the positively predicted samples across three models and the five AF datasets. Overall, ITA exhibits a beneficial impact in most scenarios, with only two out of thirty evaluations showing a marginal decline in performance.

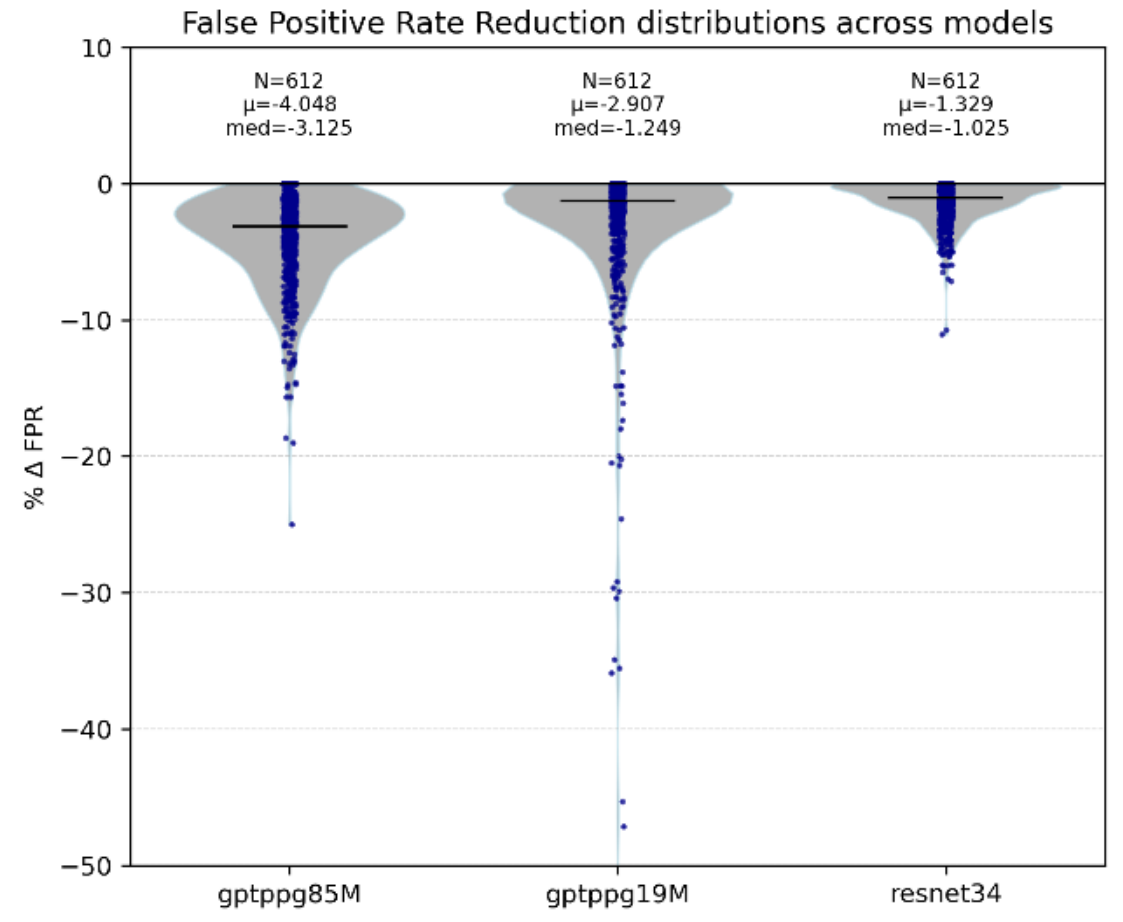

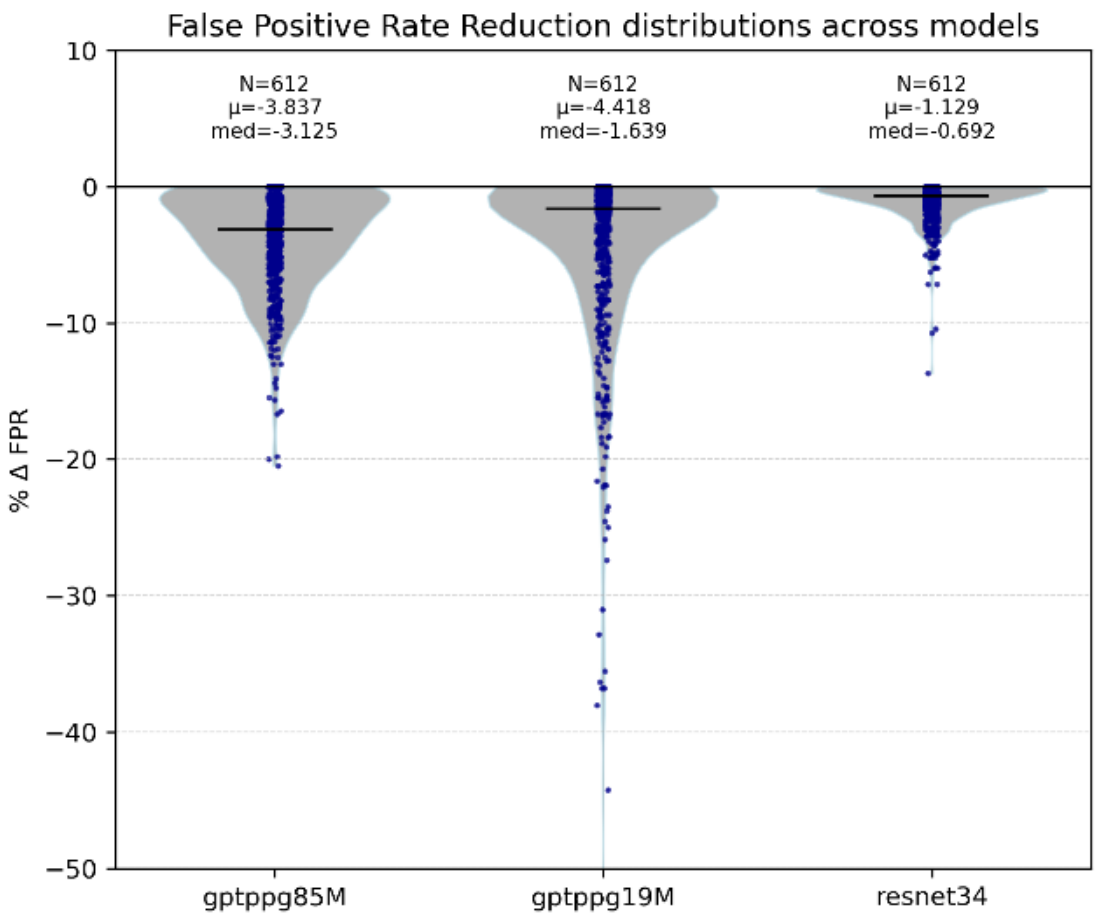

**Figure 9.** FPR reduction caused by applying ITA on positive predicted samples, represented by violine graphs for three models. Left: using the hyperparameters optimized for AUROC metric, Right: using the hyperparameters optimized for AUPRC metric. N: number of subjects, μ: mean, med: median.

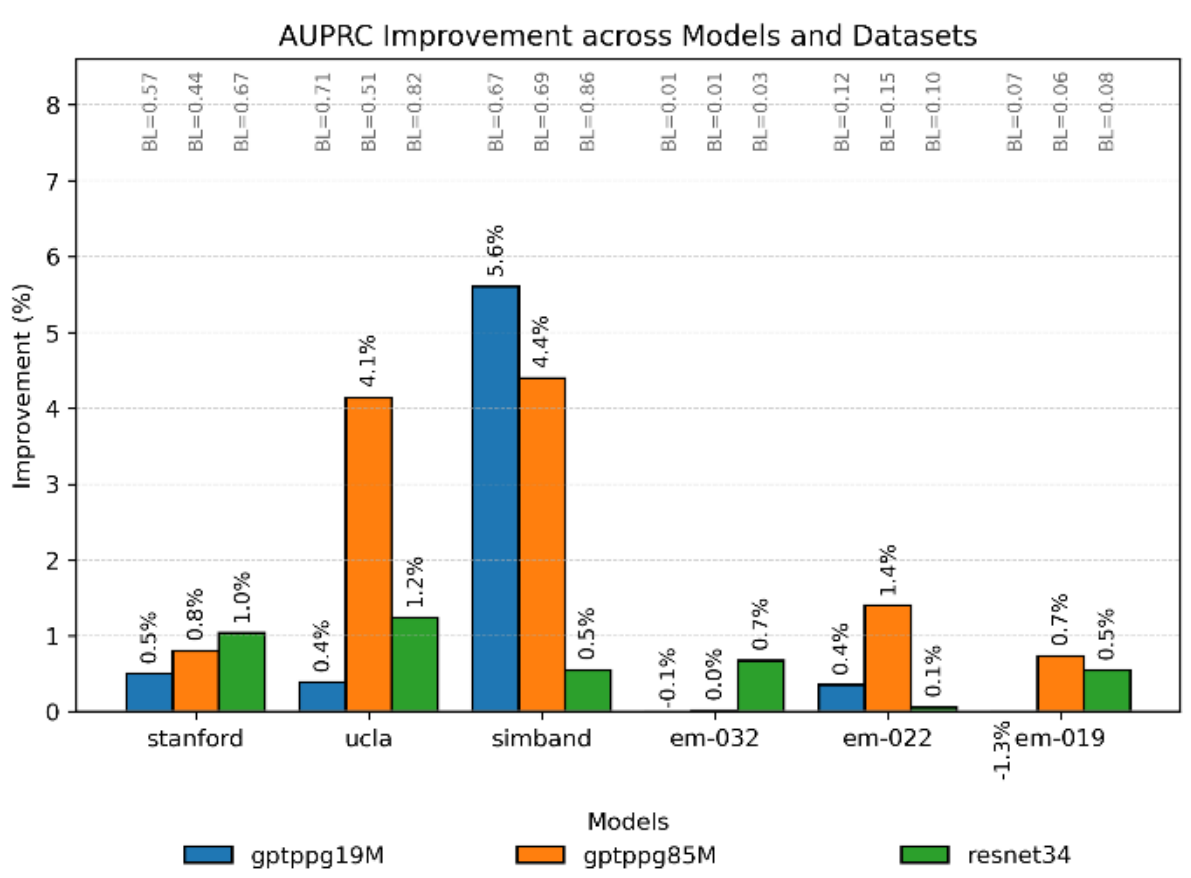

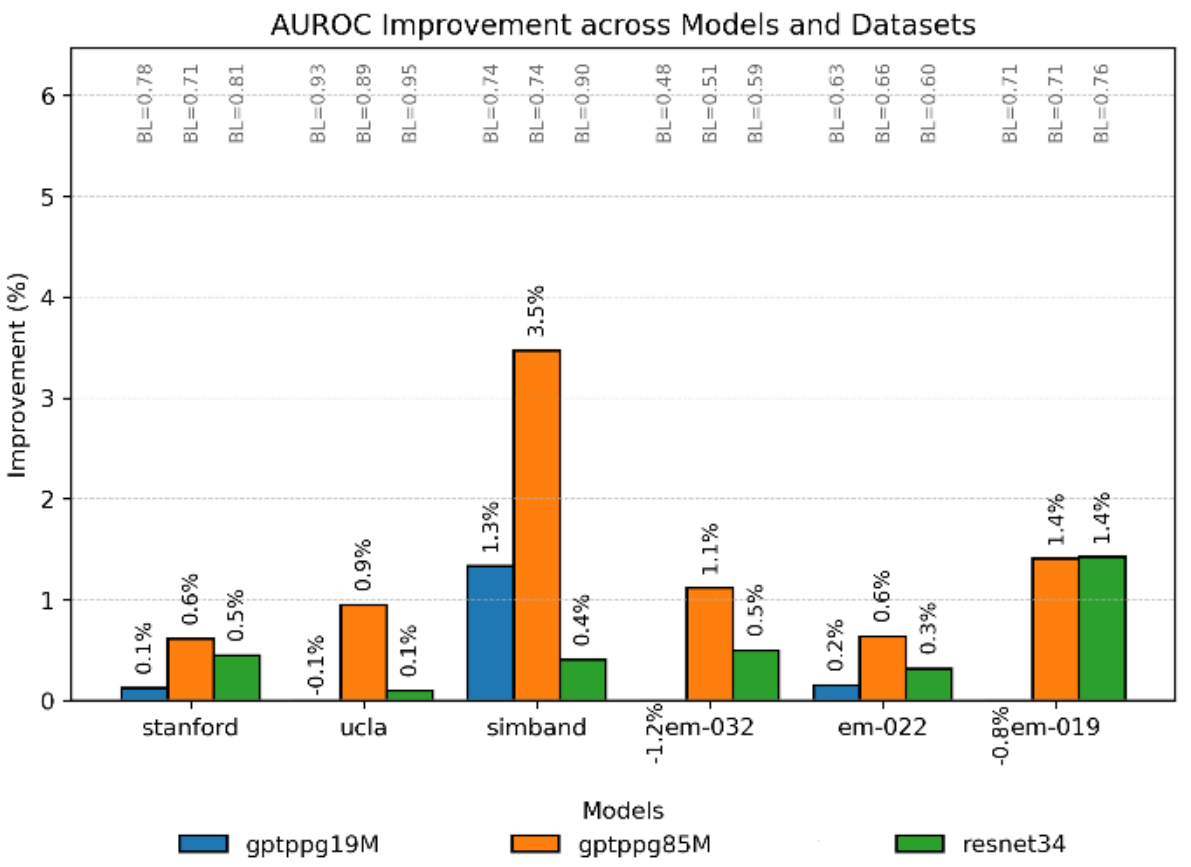

**Figure 10.** Percent of AUROC and AUROC improvement after applying ITA on positive predicted samples, for GPT-PPG and ResNet34 models, across different datasets. BL: the metric value for the baseline model (without ITA).

### *5.3 Optimized Parameters and Practical Rules of Thumb*

Table 2 summarizes the optimized augmentation parameters across different model architectures and performance metrics. Models with varying capacities and internal representations may differ in how they separate the two classes within their latent spaces; therefore, differences in optimized augmentation parameters across models are expected. However, as shown in Table 2, several parameters remain consistent, or vary only slightly, across models and evaluation metrics. This section explores these consistencies to derive practical rules of thumb, providing general guidelines that may inform future model design and augmentation tuning.

- Average number of augmented variants ranges from 4 to 7, indicating that ITA effectiveness can be achieved with a relatively few variants, making it suitable for settings with limited computational resources.
- Among different noise colors, the violet noise has been the most effective one. Although the color of noise is also affected by the subsequent frequency filtering, the violet color was the most effective one across all models and metrics. Justifiably, violet-noise ITA can improve PPG classification by perturbing each sample along high-frequency directions that do not distort physiological morphology. These perturbations expose instability of the classifier around non-robust features and allow Monte-Carlo smoothing of the decision function, yielding predictions that are more stable and consistent with low-frequency pulse morphology.
- The empirical distribution of the signal consistently emerged as the most effective noise distribution across all evaluated model–metric pairs. This indicates that noise sampled directly from the real physiological signal, rather than from theoretical distributions such as Gaussian, uniform, or Laplace, captures the nuanced variability and measurement imperfections inherent to the dataset. As a result, empirical noise provides more realistic perturbations during ITA, leading to more robust and reliable predictions.
- The optimal SNR levels are approximately 30 dB for three model–metric pairs and about 17 dB for another, correspond to noise standard deviations of roughly 3% and 14% of the signal amplitude, respectively. This highlights the sensitivity of ITA to noise intensity and suggests that low noise levels, far below the signal scale, are most effective for improving prediction robustness.
- Sinusoidal interference shows limited contribution to the ITA performance, evidenced by near-zero optimized amplitudes in three model-metric combinations. In the other cases, the sinusoidal frequency converges to a high frequency near the Nyquist limit, which is consistent with the fact that such frequencies lie outside low-frequency band of PPG's informative morphology. A plausible explanation is that the PPG signals in both pretraining and HPO phases were already band-limited to approximately 20 Hz, effectively removing powerline noise and other sinusoidal interferences. As a result, adding such components during ITA provides little benefit, since the model has not been exposed to these types of distortions.
- Basic deterministic augmentations, including time flipping, amplitude inversion, and linear amplitude drift, did not improve performance in any of the model–metric pairs. This lack of effectiveness is likely due to the absence of tunable parameters in these methods, which results in overly drastic transformations of the signal and the loss of physiologically informative PPG characteristics.

## 6. Conclusion

In this study, we introduced a comprehensive ITA framework for physiological signals that integrates a broad set of augmentation methods across time, amplitude, frequency, artifact, and noise domains. The framework unifies these transformations under a consistent parameterization with 39 tuneable hyperparameters, enabling flexible and metric-specific customization without retraining the underlying model. Using Bayesian optimization, we identified optimal augmentation configurations and explored their impact across multiple performance metrics.

We demonstrated the effectiveness of the proposed framework on the task of PPG-based AF detection using two DNN models. ITA consistently improved AUROC, and AUPRC across all models and datasets. We further introduced a selective ITA strategy that applies augmentation only to positively predicted samples, yielding reductions in false positives. Latent-space visualizations confirmed that ITA produces more stable and separable class representations, reinforcing its role as a robustification mechanism at inference time.

While augmentation methods have been extensively discussed in the literature of AI, the present study can contribute to this field by the following innovations.

First, we present a new unified ITA framework tailored for physiological signals, offering a systematic design that comprehensively spans multiple augmentation domains with extensive tunability. In addition, in an innovative way, we introduce Bézier and PCHIP curves as general nonlinear operators for time-warping, amplitude modulation, and amplitude drift, producing smooth, and length-preserving transformation which is compatible with physiological signals augmentation.

**Table 2.** Optimized hyper parameters of ITA frameworks

| **Method** | Parameters | HPO results for AUROC metric | | | HPO results for AUPRC metric | | |
|---|---|---|---|---|---|---|---|
| | | **GPT-PPG 85M** | **GPT-PPG 19M** | **Resnet34** | **GPT-PPG 85M** | **GPT-PPG 19M** | **Resnet34** |
| Fused augmentation | Average No. of aug. variants | 6 | 6 | 7 | 7 | 5 | 4 |
| Additive Noise | Signal-to-noise ratio (dB) | 30.0 | 30.0 | 30.0 | 30.0 | 30.0 | 16.85 |
| | Noise color | violet | violet | violet | violet | violet | violet |
| | Bandpass filter low cutoff | 0.18 | 0.4 | 0.0 | 0.30 | 0.0 | 0.16 |
| | Bandpass filter high cutoff | 0.97 | 0.5 | 0.5 | 0.81 | 0.5 | 0.51 |
| | Noise distribution | resample | resample | resample | resample | resample | resample |
| Bézier-Based Time Warping | Warping intensity | 0.100 | 0.001 | 0.001 | 0.100 | 0.002 | - |
| | Number of control knots | 3 | 3 | 3 | 3 | 3 | - |
| Bézier-Based Amplitude Drift | Drift intensity | 0.06 | 0.035 | 0.044 | 0.088 | 0.035 | 0.088 |
| | Number of control knots | 4 | 5 | 5 | 6 | 5 | 5 |
| Bézier-Based Amp. Modulation | Modulation intensity | 0.114 | 0.070 | 0.107 | 0.130 | 0.055 | 0.156 |
| | Number of control knots | 5 | 3 | 4 | 4 | 4 | 5 |
| PCHIP-Based Time Warping | Warping intensity | 0.017 | 0.015 | 0.002 | 0.004 | 0.004 | 0.002 |
| | Number of control knots | 3 | 3 | 3 | 5 | 5 | 3 |
| PCHIP-Based Amplitude Drift | Drift intensity | 0.047 | 0.001 | 0.013 | 0.047 | 0.047 | 0.049 |
| | Number of control knots | 3 | 6 | 6 | 3 | 3 | 5 |
| PCHIP-Based Amp. Modulation | Modulation intensity | 0.069 | 0.028 | 0.055 | 0.086 | 0.047 | 0.068 |
| | Number of control knots | 4 | 5 | 4 | 5 | 5 | 4 |
| Low-Frequency Noise Injection | Noise amplitude | 0.010 | 0.016 | 0.016 | 0.016 | 0.016 | 0.016 |
| | Low-pass cut-off freq. | 0.010 | 0.048 | 0.048 | 0.048 | 0.048 | 0.048 |
| Sinusoidal Interference | Sinusoidal wave amplitude | 0.013 | 0.048 | - | - | 0.013 | 0.0002 |
| | Sinusoidal frequency | 0.095 | 0.095 | - | - | 0.095 | 0.021 |
| Burst Artifact Injection | Burst amplitude | 0.678 | 0.070 | 0.385 | 0.71 | 0.260 | 0.350 |
| | No. of bursts per signal | 4 | 3 | 1 | 8 | 1 | 7 |
| | Burst noise frequency | 0.01 | 0.01 | 0.01 | 0.01 | 0.01 | 0.01 |
| Basic Deterministic Transformations | Flip in time | false | false | false | false | false | false |
| | Invert amplitude | false | false | false | false | false | false |
| | Apply linear baseline drift | false | false | false | false | false | false |

We also generalize conventional noise-augmentation techniques into a flexible and highly configurable module capable of incorporating multiple noise colors, distributions, filtering schemes, and SNR ranges. Generalization and parametrization are also applied for other augmentation techniques like artifact injections. Furthermore, we develop a scalable Bayesian hyperparameter optimization scheme compatible with large search. Also, we validate the idea of selective ITA as an inference strategy to reduce false positives. Finally, our experiments provide empirical insights into augmentation effectiveness, highlighting some rules of thumb for leveraging ITA in PPG AF detection.

Overall, our findings establish ITA as a practical and effective approach for enhancing the robustness of physiological signal classification. The framework is general, extensible, and readily applicable to other physiological modalities and downstream clinical tasks.

Several promising research directions emerge from this study. For instance, an important open question is determining when ITA should be applied during deployment. While this work investigated both full-sample and selective ITA strategies, future research could develop confidence-aware or entropy-based decision rules to automatically decide whether a given input requires augmentation, balancing computational cost with performance gains.

In addition, although this study focused on PPG signals, the proposed framework is general and can be extended to other physiological modalities, such as ECG, PCG, EMG, EEG, and EOG. Each modality may benefit from modality-specific augmentation primitives or constraints, and evaluating ITA across these domains could establish broader applicability in clinical and wearable settings.

Furthermore, extending ITA beyond classification to other problem types, including regression (e.g., estimating hemodynamic parameters), segmentation, or anomaly detection, represents another avenue for advancement. Tailoring augmentation strategies and aggregation functions for regression tasks, for example, through noise-aware averaging or uncertainty-weighted ensembles, could enable ITA to improve performance across a wider range of physiological signal analysis applications.

Finally, integrating ITA with other augmentation schemes like contrastive learning could strengthen representation robustness by encouraging the model to learn invariant features across augmented views of the same physiological signal. Unsupervised contrastive learning represents a particularly promising extension of this work. Because ITA relies on generating diverse yet physiologically meaningful signal variants, these same augmentations can serve as positive pairs for contrastive pretraining. Given the high label noise and substantial inter-individual variability in physiological datasets (especially PPG), contrastive learning could substantially improve representation robustness and, in turn, further enhance the benefits of ITA in downstream tasks such as AF detection.

## Data and Code Availability

The augmentation package developed in this work is publicly available on GitHub (https://github.com/davood-fattahi/augsig), and on PyPI (https://pypi.org/project/augsig/).

The Python implementation of the DA-PCA projection and related visualization tools introduced in this work is publicly available on GitHub (https://github.com/davood-fattahi/da-pca).

The external datasets used in this study are cited in the text and identified in the references; the non-AF PPG datasets and their repository links are additionally listed in Appendix A5. The Delta dataset is an ongoing NIH-funded study (grant R01HL166233); data collection is expected to be completed by approximately 2027, after which it will be made publicly available.

## Acknowledgement

This work was partially supported by the National Institutes of Health through grant numbers R01HL166233.

## Appendix

### *A.1 Adding Noise*

Introducing controlled levels of synthetic noise during augmentation can simulate these real-world distortions, enabling the model to learn to focus on the underlying physiological patterns rather than overfitting to clean, idealized inputs.

Mathematically, a noisy signal $\tilde{x}(t)$can be represented as

$$\tilde{x}(t) = x(t) + n(t), \qquad \text{Eq. (A.1)}$$

where $x(t)$ is the original physiological signal and $n(t)$ is the synthetic noise component. A key parameter is the signal-to-noise ratio (SNR), which controls the intensity of the added noise. SNR is commonly defined as

$$\mathrm{SNR}_{\mathrm{dB}} = 10\log_{10}\left(\frac{\mathbb{E}[x(t)^2]}{\mathbb{E}[n(t)^2]}\right), \qquad \text{Eq. (A.2)}$$

and the noise amplitude can be scaled to achieve a target SNR via

$$n_{\mathrm{scaled}}(t) = n_{\mathrm{raw}}(t) \cdot \sqrt{\frac{\mathbb{E}[x(t)^2]}{\mathbb{E}[n_{\mathrm{raw}}(t)^2] \cdot 10^{\frac{\mathrm{SNR_{dB}}}{10}}}}. \qquad \text{Eq. (A.3)}$$

Another factor is the type or color of noise which can be determined by its power spectral density (PSD). In order to incorporate flat, monotonically increasing, and monotonically decreasing shapes of PSD in our study, we consider the following noise colors: White noise ($S(f) \propto f^0$), Pink noise ($S(f) \propto \frac{1}{f}$), Brown noise ($S(f) \propto \frac{1}{f^2}$), Violet noise ($S(f) \propto f^2$). Additionally, the noise can also be more customized in frequency domain by applying bandpass filters. This can be expressed using a frequency-domain mask $H(f)$:

$$n_{\mathrm{filtered}}(t) = \mathcal{F}^{-1}[H(f) \cdot \mathcal{F}\{n(t)\}], \qquad \text{Eq. (A.4)}$$

where $H(f)$ may represent low-pass, high-pass, or band-pass filters depending on the intended noise profile.

The distribution of noise samples also affects realism. In the present study we consider four different distributions: Gaussian noise $n(t) \sim \mathcal{N}(0, \sigma^2)$, uniform noise $n(t) \sim \mathcal{U}(-a, a)$, Laplace noise $n(t) \sim \mathrm{Laplace}(0, b)$, and resampled from the empirical distribution of the signal $n(t) \sim \mathrm{Empirical}(x)$.

### *A.2 Adding Burst*

Burst noise augmentation introduces short, high-intensity disturbances into the signal to simulate transient artifacts that commonly occur in real-world physiological recordings. These bursts may result from sudden sensor movements, brief loss of contact, muscle twitches, or abrupt environmental interferences. In biomedical signals such as PPG or ECG, such artifacts can significantly distort a few consecutive samples while leaving the rest of the signal relatively unaffected. By adding controlled bursts during augmentation, the model learns to tolerate and correctly interpret signals containing localized disruptions. In our implementation, burst noise is defined by parameters such as amplitude, number of bursts, and frequency of occurrence, allowing precise control over their strength and distribution. This approach helps ensure that models remain robust to sporadic, high-energy artifacts that may otherwise lead to misclassification or unstable predictions in deployment.

### *A.3 Adding Sine Wave*

Sine wave noise augmentation involves adding a sinusoidal signal to the physiological waveform to mimic periodic interference commonly encountered in real-world measurements. Such interference can arise from sources like powerline contamination (e.g., 50/60 Hz mains hum), mechanical oscillations from equipment, or cyclic physiological processes unrelated to the target measurement.

Incorporating sine wave noise in augmentation allows the model to learn to recognize and ignore predictable periodic disturbances while focusing on the underlying diagnostic patterns. In our framework, the sine wave is characterized by parameters such as amplitude and frequency, enabling the generation of low-frequency drifts, mid-frequency respiratory-like oscillations, or high-frequency powerline-like noise. By varying these parameters, we can expose the model to a wide range of periodic interference scenarios, enhancing its resilience to common environmental and instrumental artifacts.

### *A.4 Flip and Invert*

Flip and invert augmentations modify the polarity or temporal orientation of a physiological signal, introducing variations that help the model become invariant to changes in signal presentation. Flipping reverses the signal along the time axis, simulating scenarios where the acquisition device outputs time-reversed data due to configuration differences or signal processing pipelines. Inverting multiplies the signal by −1, reversing its vertical polarity, which can occur in practice due to electrode lead reversal, sensor placement differences, or hardware-specific output conventions. While these transformations preserve the essential morphological features of the waveform, they alter its orientation or direction, forcing the model to learn representations that are robust to such variations. In our implementation, both operations are applied independently or in combination, ensuring that the model is not overly sensitive to polarity or temporal orientation changes in physiological signals.

Figure A.1 illustrates examples for different augmentation methods applied on a 30second PPG signal.

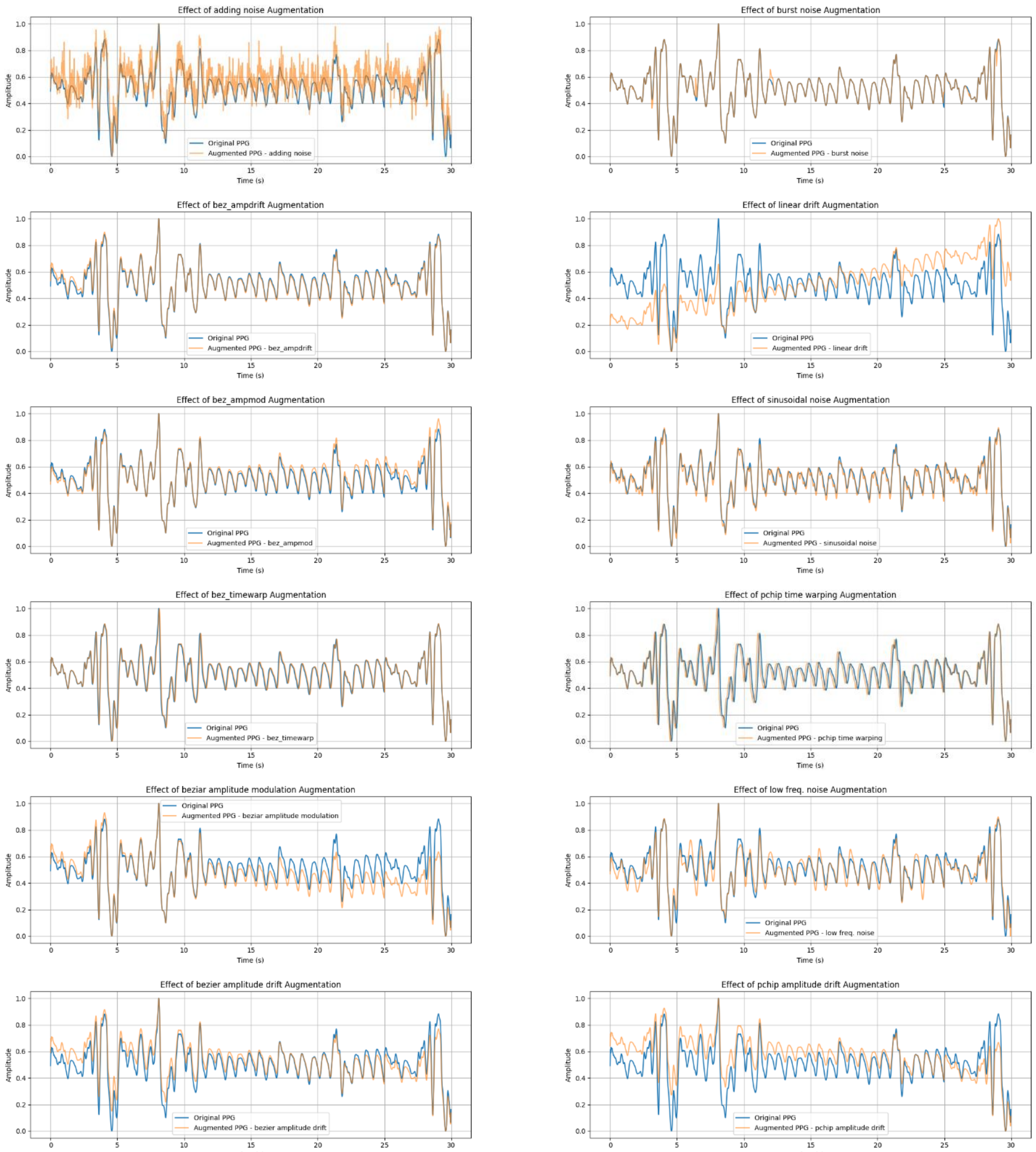

**Figure A.1.** Examples for different augmentation methods applied on a 30second PPG signal.

## *A.5 List of Non-AF PPG Datasets*

Table A.1. List of non-AF PPG datasets

| No. | Title | Ref. | Dataset Address |
|---|---|---|---|
| 1 | WildPPG: A Real-World PPG Dataset of Long Continuous Recordings | [37] | https://siplab.org/projects/WildPPG |
| 2 | GalaxyPPG: A PPG Signal Dataset Collected in Semi-Naturalistic Settings Using Galaxy Watch | [38] | https://zenodo.org/records/14635823 |
| 3 | Development of a Screening Tool for Sleep Disordered Breathing in Children Using the Phone Oximeter | [39] | https://figshare.com/articles/dataset/Development_of_a_Screening_Tool_for_Sleep_Disordered_Breathing_in_Children_Using_the_Phone_Oximeter/1209662/6 |
| 4 | ECSMP: A Dataset on Emotion, Cognition, Sleep, and Multi-model Physiological Signals | [40] | https://data.mendeley.com/datasets/vn5nknh3mn/2 |
| 5 | Motion Artifact Cancellation in Wearable Photoplethysmography Using Gyroscope | [41] | https://github.com/hooseok/gyro_acc_ppg |
| 6 | Pulse Transit Time PPG Dataset | [42] | https://physionet.org/content/pulse-transit-time-ppg/1.0.0/ |
| 7 | PPG-DaLiA | [43] | https://archive.ics.uci.edu/dataset/495/ppg+dalia |
| 8 | WESAD (Wearable Stress and Affect Detection) | [44] | https://archive.ics.uci.edu/dataset/465/wesad+wearable+stress+and+affect+detection |
| 9 | iAMwell: ECG, PPG, RSP signals from athletes and non-athletes during exercise | [45] | https://zenodo.org/records/1012726 |
| 10 | Simultaneous physiological measurements with five devices at different cognitive and physical loads | [46] | https://physionet.org/content/simultaneous-measurements/1.0.0/ |
| 11 | AffectiveROAD system and database to assess driver's attention | [47] | https://www.media.mit.edu/groups/affective-computing/data/ |
| 12 | WRIST: Wrist PPG During Exercise | [48] | https://physionet.org/content/wrist/1.0.0/ |
| 13 | Dataset from PPG wireless sensor for activity monitoring | [49] | https://doi.org/10.1016/j.dib.2019.105044 |
| 14 | Raw PPG Signal Measured Using Wearable Sensor-kit in Varying Levels of Activity | [50] | https://purr.purdue.edu/publications/3180/1 |
| 15 | PPG Diary 1 Study Data | [51] | https://zenodo.org/records/5211472 |
| 16 | Eight-Emotion Sentics Data | [52] | https://www.media.mit.edu/groups/affective-computing/data/ |
| 17 | OpenOximetry Repository | [53] | https://physionet.org/content/openox-repo/1.0.0/ https://openoximetry.org/data-repository/ |

## *A.6 Theoretical Basis of Decision-Aligned PCA*

### *A.6.1 Problem Setup*

Let $z \in \mathbb{R}^d$ denote a feature vector and let

$$f(z) = w^T z + b \qquad \text{Eq. (A.5)}$$

be a linear classifier with weight vector $w \in \mathbb{R}^d$ and bias $b \in \mathbb{R}$. The signed distance from $z$ to the decision hyperplane $\{z : f(z) = 0\}$ is

$$d(z) = \frac{w^T z + b}{\|w\|}. \qquad \text{Eq. (A.6)}$$

Given a training set $\{z_i\}_{i=1}^N \subset \mathbb{R}^d$, DA-PCA constructs an orthonormal basis $(a_1, a_2)$ and a center $\mu$ such that projecting onto this basis yields a 2D representation that is both margin-preserving and data-informed.

### *A.6.2 Construction of the Projection Basis*

*Step 1- Decision axis:* Define the unit vector aligned with the classifier's decision normal:

$$a_1 = \frac{w}{\|w\|}. \qquad \text{Eq. (A.7)}$$

*Step 2- Center:* Compute the mean of the training features:

$$\mu = \frac{1}{N}\sum_{i=1}^{N} z_i. \qquad \text{Eq. (A.8)}$$

*Step 3- Residuals:* Remove the $a_1$ component from each centered point:

$$r_i = (z_i - \mu) - [(z_i - \mu)^T a_1]a_1. \qquad \text{Eq. (A.9)}$$

*Step 4- Spread axis:* Let $a_2$ be the first principal component of the residual matrix $R = [r_1, \dots, r_N]^T$, i.e., the unit vector solving

$$a_2 = \underset{\|v\|=1, v \perp a_1}{\operatorname{argmax}} \; v^T R^T R v. \qquad \text{Eq. (A.10)}$$

By construction, $a_2 \perp a_1$ and $\|a_2\| = 1$.

*Step 5- Projection:* Any feature vector $z$ is mapped to 2D coordinates $(x_1, x_2)$ by

$$x_1 = {a_1}^T (z - \mu), \quad x_2 = {a_2}^T (z - \mu). \qquad \text{Eq. (A.11)}$$

### *A.6.3 Margin-Preservation Property*

*Proposition:* The first coordinate $x_1$ equals the true signed margin $d(z)$ up to a constant shift:

$$x_1 = d(z) - d(\mu). \quad \text{Eq. (A.12)}$$

*Proof:* Substituting $a_1 = \frac{w}{||w||}$:

$$x_1 = {a_1}^T(z - \mu) = \frac{w^T z}{||w||} - \frac{w^T \mu}{||w||}$$
$$= \frac{w^T z + b}{||w||} - \frac{w^T \mu + b}{||w||}$$
$$= d(z) - d(\mu). \quad \text{Eq. (A.13)}$$

The centering constant $d(\mu)$ is fixed for a given training set, so differences in $x_1$ between any two points exactly equal differences in their signed margins. Moving right in the 2D plane increases the classifier score; moving left decreases it.

### *A.6.4 Decision Boundary in 2D*

*Proposition:* The decision hyperplane $f(z) = 0$ projects to a vertical line in the DA-PCA plane at

$$x_1 = c, \quad c = -\frac{w^T \mu + b}{||w||}. \quad \text{Eq. (A.14)}$$

*Proof:* Substitute $z = \mu + x_1\, a_1 + x_2\, a_2$ into$f(z) = 0$:

$$w^T \mu + (w^T a_1)x_1 + (w^T a_2)x_2 + b = 0 \quad \text{Eq. (A.15)}$$

Since $a_1 = \frac{w}{||w||}$, we have $w^T a_1 = ||w||$. Since $a_2 \perp a_1$ and $w \parallel a_1$, we have $w^T a_2 = 0$. Therefore:

$$||w||x_1 + w^T \mu + b = 0$$
$$\Longrightarrow x_1 = -\frac{w^T \mu + b}{||w||} = c. \quad \text{Eq. (A.16)}$$

The absence of an $x_2$ term confirms that the boundary is exactly vertical regardless of the data distribution. In general form, the projected boundary satisfies $Ax_1 + Bx_2 + C = 0$ with $A = ||w||$, $B = 0$, and $C = w^T \mu + b$.

### *A.6.5 Relationship to Standard PCA*

Standard PCA selects the 2D subspace of maximum total variance. DA-PCA fixes one axis to $a_1 = \frac{w}{||w||}$ and maximizes residual variance for the second axis. These two methods are generally distinct.

*Proposition:* DA-PCA is not obtainable by an in-plane rotation of standard PCA unless $w$ lies entirely within the top-2 PCA subspace.

*Justification:* The top-2 PCA subspace $V_{PCA}$ spans the two maximum-variance directions of the data covariance $\Sigma$. The DA-PCA subspace $V_{DA}$ is spanned by $a_1 = w/||w||$ and the leading eigenvector of the residual covariance in $a_1 \perp$. If $w \notin V_{PCA}$ (which holds generically when $w$ has nonzero projection onto low-variance directions of $\Sigma$) then $V_{DA} \neq V_{PCA}$, and no rotation within $V_{PCA}$ can produce $a_1$. Table A.2 summarizes the key differences.

Table A.2. Comparing DA-PCA with Standard PCA

| Property | Standard PCA | DA-PCA |
|---|---|---|
| **First axis** | Maximum-variance direction | $w/\|\|w\|\|$ (classifier normal) |
| **Second axis** | Maximum variance orthogonal to first | Maximum variance orthogonal to first |
| **Depends on classifier** | No | Yes |
| **Globally variance-optimal** | Yes | No (conditional) |
| **Margin directly readable** | No | Yes ($x_1 = d(z) - d(\mu)$) |
| **Boundary always vertical** | No | Yes |

### *A.6.6 Train/Test Discipline*

The basis ($a_1$, $a_2$, $\mu$) must be fit on training data only and frozen for all subsequent projections (validation, test, augmented inputs). This is essential for two reasons: $a_1 = \frac{w}{||w||}$ derives from the trained classifier, which was fit on training data. Fitting $a_2$ or $\mu$ on test data leaks distributional information into the projection frame, subtly shifting the apparent position of the decision boundary and making any observed displacement of test-time features an artifact of the frame rather than a genuine geometric movement.

With a fixed training-fit basis, any shift in $x_1$ between two conditions (e.g., baseline versus augmented features) reflects a genuine change in signed margin relative to the training decision geometry.